%% file: arxiv.tex
\documentclass[10pt,twocolumn,letterpaper]{article}
\pdfoutput=1 

\usepackage{cvpr}
\usepackage{amsmath,amssymb,amsfonts,gensymb}

\cvprfinalcopy
\usepackage{tikz}
\usetikzlibrary{decorations.pathmorphing}
\usetikzlibrary{intersections}
\usetikzlibrary{positioning}
\usetikzlibrary{calc}
\usetikzlibrary{fit}
\usepackage[skins]{tcolorbox}

\usepackage{times}
\usepackage{graphicx}
\usepackage{adjustbox}
\usepackage{array}
\def\rot{\rotatebox}
\usepackage{tabu}

\usepackage{upgreek}
\usepackage{bm}
\usepackage{comment}
\makeatletter
\newcommand{\bfgreek}[1]{\bm{\@nameuse{#1}}}
\newcommand{\bfgreekco}[2]{\bm{\textcolor{#2}{\@nameuse{#1}}}}
\makeatother

\usepackage[pagebackref=true,breaklinks=true,letterpaper=true,colorlinks,bookmarks=false]{hyperref}

\begin{document}

\title{Feedforward semantic segmentation with zoom-out features}

\makeatletter
\author{Mohammadreza Mostajabi, Payman Yadollahpour and Gregory Shakhnarovich\\
Toyota Technological Institute at Chicago\\{\tt \{mostajabi,pyadolla,greg\}@ttic.edu}}
\makeatother

\date{}

\maketitle

\begin{abstract}
We introduce a purely feed-forward architecture for semantic
segmentation. We map small image elements (superpixels) to rich
feature representations extracted from a sequence of nested regions of
increasing extent. These regions are obtained by "zooming out" from the superpixel all
the way to scene-level resolution. This approach exploits
statistical structure in the image and in the label space without setting up explicit structured
prediction mechanisms, and thus avoids complex and expensive
inference. Instead superpixels are classified by a feedforward multilayer network. Our architecture achieves new
state of the art performance in semantic segmentation,
obtaining \textbf{64.4\%} average accuracy on the PASCAL VOC 2012 test set.
\end{abstract}


\input{intro}

\input{method}

\input{related}

\input{experiments}

\input{conclusions}

{\small
\bibliographystyle{ieee}
\bibliography{segmentation}
}

\end{document}

%% file: intro.tex
\section{Introduction}
We consider one of the central vision tasks, \emph{semantic
  segmentation}: assigning to each
pixel in an image a category-level label. Despite attention it has
received, it remains challenging, largely due to complex
interactions between neighboring as well as distant image elements,
the importance of global context, and the interplay between semantic
labeling and instance-level detection.
A widely accepted conventional wisdom, followed in much of
modern segmentation literature, is that 
segmentation should be treated as a structured prediction task, which most
often means using a random field or structured support vector machine
model of considerable complexity.

This in turn brings up severe challenges, among them the intractable
nature of inference and learning in many ``interesting'' models. To alleviate this,
many recently proposed methods rely on a pre-processing
stage, or a few stages, to produce a manageable number of hypothesized
regions, or even complete segmentations, for an image. These are then
scored, ranked or combined in a variety of ways.

\input{mainfig}

Here we consider a departure from these conventions, and approach semantic
segmentation as a single-stage classification task, in which each
image element (superpixel) is labeled by a feedforward model, based on evidence
computed from the image. Surprisingly, in experiments on PASCAL VOC 2012
segmentation benchmark we show that this simple sounding approach
leads to results significantly surpassing all previously published
ones, advancing the current state of the
art from about 52\% to 64.4\%.

The ``secret'' behind our method is that the evidence used in the
feedforward classification is not computed from a small local region
in isolation, but collected from a sequence of levels, obtained by
``zooming out'' from the close-up view of the superpixel. Starting from the superpixel itself, to a small region surrounding it,
to a larger region around it and all the way to the entire image, we
compute a rich feature representation at each level and combine all
the features before feeding them to a classifier. This allows us to
exploit statistical structure in the label space and dependencies
between image elements at different resolutions without explicitly
encoding these in a complex model.

We should emphasize that we do not mean to dismiss structured
prediction or inference, and as we discuss in
Section~\ref{sec:conclusions}, these tools may be complementary to our
architecture. In this paper we explore how far we can go
without resorting to explicitly structured models.

We use convolutional neural networks (convnets) to extract features from
larger zoom-out regions. Convnets, (re)introduced
to vision in 2012, have facilitated a dramatic advance in classification,
detection, fine-grained recognition and other vision tasks. Segmentation has remained conspicuously left out from this wave of
progress; while image classification and detection accuracies on
VOC have improved by nearly 50\% (relative), segmentation numbers have
improved only modestly. A big reason for this is that neural networks
are inherently geared for ``non-structured'' classification and
regression, and it is still not clear how they can be harnessed in a
structured prediction framework. In this work we propose a way to
leverage the power of representations learned by convnets, by framing
segmentation as classification and making the structured aspect of it
implicit. 
Last but not least, we show that use of
multi-layer neural network trained with asymmetric loss to
classify superpixels represented by zoom-out features, leads
to significant improvement in segmentation accuracy over simpler
models and conventional (symmetric) loss.

Below we give a high-level description of our method, then discuss
related work and position our work in its context. Most of the
technical details are deferred to Section~\ref{sec:exp} in which we
describe implementation and report on results, before concluding in Section~\ref{sec:conclusions}.

%% file: mainfig.tex
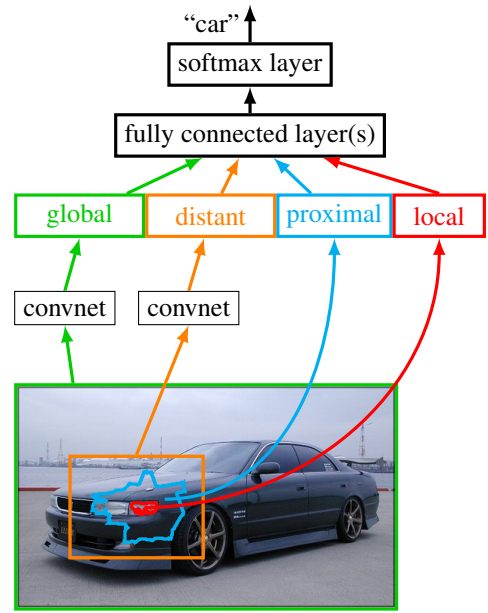
\begin{figure}[!th]
\centering
  \begin{tikzpicture}
    \node[rectangle,draw=black,thick,fill overzoom
    image=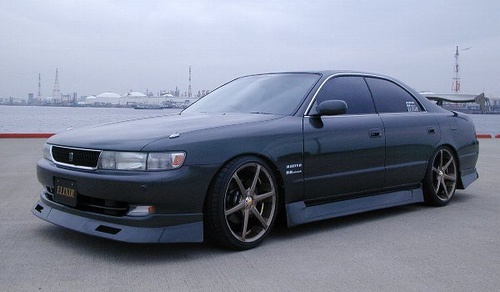,minimum width=5cm,minimum height=2.92cm,opacity=.5] (image) {};

    \draw[name path=superpixel,red,line width=1.25pt] plot[smooth] file {figs/2007_003143_s252_n0.txt};
    \coordinate (suppixcenter) at (-.82,-.14);

    \draw[name path=n2,cyan,line width=1.25pt] plot file
    {figs/2007_003143_s252_n2.txt};
    \coordinate (n2pt) at (-.55,-.03);
    
    \coordinate (n3bl) at (-1.81,-.81);
    \coordinate (n3tr) at (-.04,.54);
    \node[draw,orange,line width=1.25pt,inner sep=0pt,fit=(n3bl) (n3tr)] (n3) {};
    \node[draw,very thick,black!20!green,fit=(image),inner sep=0.5pt] (global) {};
    
    \node[draw,black,above=of global.north west,anchor=west]
    (dcnn-global) {convnet};
    \node[draw,black!20!green,above=of dcnn-global.north
    west,anchor=west,minimum height=5.5mm,minimum width=1.7cm,very thick]
    (fglobal) {global};
    \draw[black!20!green,-latex,very thick] ($(global.130)-(.5,0)$) -- (dcnn-global.south);
    \draw[black!20!green,-latex,very thick] (dcnn-global.north) --
    (fglobal.south);

    \node[draw,black,right=of dcnn-global.east,right=3mm,anchor=west]
    (dcnn-n3) {convnet};
    \node[draw,orange,right=of
    fglobal.east,anchor=west,right=0pt,minimum height=5.5mm,minimum
    width=1.7cm,very thick]
    (fn3) {distant};
    \draw[orange,-latex,very thick] (n3.north) -- (dcnn-n3.south);
    \draw[orange,-latex,very thick] (dcnn-n3.north) -- (fn3);
    (fglobal.south);

    \node[draw,cyan,right=of
    fn3.east,anchor=west,right=0pt,minimum height=5.5mm,minimum
    width=1.2cm,very thick]
    (fn2) {proximal};
    \path[name path=n2line] (suppixcenter) -- (image.north
    east);
    \draw[-latex,cyan,very thick] (n2pt) to[out=0,in=-90] (fn2.south);

    \node[draw,red,right=of
    fn2.east,anchor=west,right=0pt,minimum height=5.5mm,minimum
    width=1.2cm,very thick]
    (flocal) {local};
    \draw[-latex,red,very thick] (suppixcenter) to[out=0,in=-90]  (flocal.south);

    \node[fit=(fglobal) (flocal)] (allfeat) {};
    \node[draw,black,very thick,above=of allfeat.north,anchor=south,above=10pt]
    (hidden) {fully connected layer(s)};
    \draw[black!20!green,-latex,very thick] (fglobal) -- (hidden);
    \draw[orange,-latex,very thick] (fn3) -- (hidden);
    \draw[-latex,cyan,very thick] (fn2) -- (hidden);
    \draw[-latex,red,very thick] (flocal.north) -- (hidden);
    \node[draw,black,very thick,above=of hidden.north,anchor=south,above=10pt]
    (softmax) {softmax layer};
    \draw[black,very thick,-latex] (hidden) -- (softmax);
    \draw[black,very thick,-latex] (softmax) -- ++(0,.8) node[left,midway] {``car''};

  \end{tikzpicture}
  \caption{Our feedforward segmentation process. The feature vector for a
    superpixel consists of components extracted at
  zoom-out spatial levels: locally at a superpixel (red), in a small
  proximal neighborhood
  (cyan), in a larger distant neighborhood (orange), and globally from the
  entire image (green). The concatenated feature vector is fed to a
  multi-layer neural network that classifies the superpixel.}\label{fig:main}
\end{figure}

%% file: method.tex
\section{Zoom-out feature fusion}\label{sec:method}
\begin{figure*}[!th]
  \centering
  \begin{tabular}{ccc}
  \includegraphics[height=1.8in]{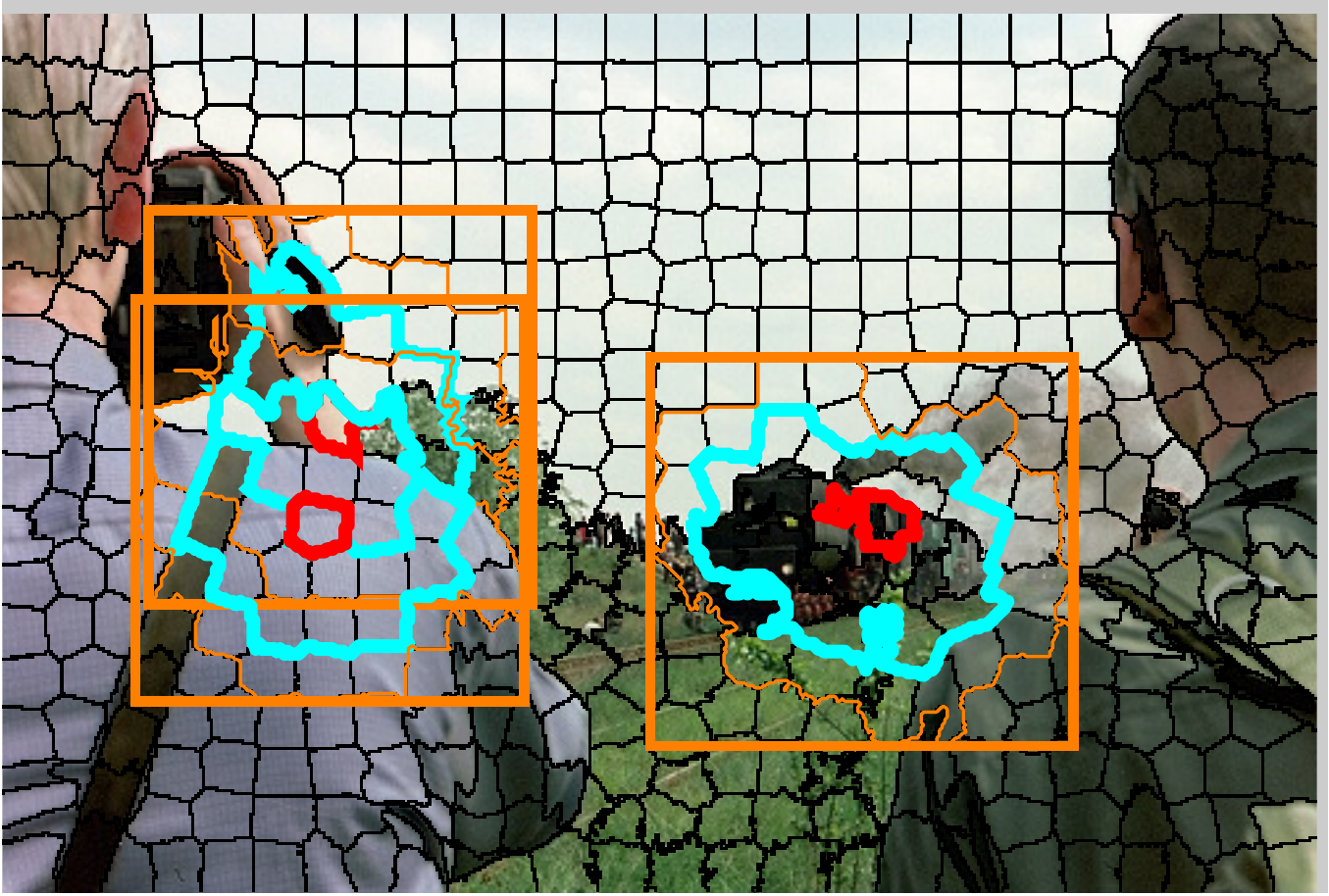} &
  \includegraphics[height=1.8in]{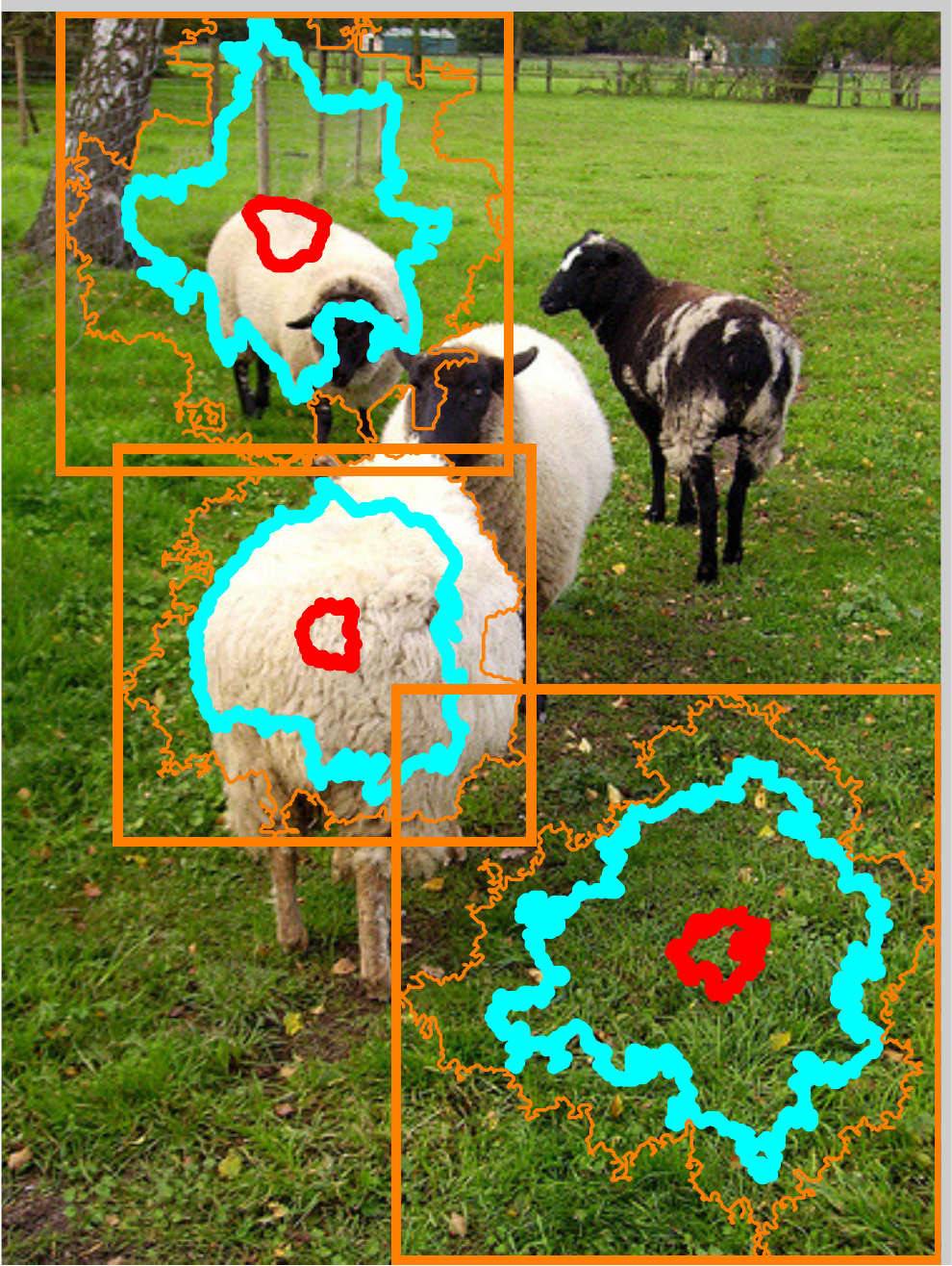} &
  \includegraphics[height=1.8in]{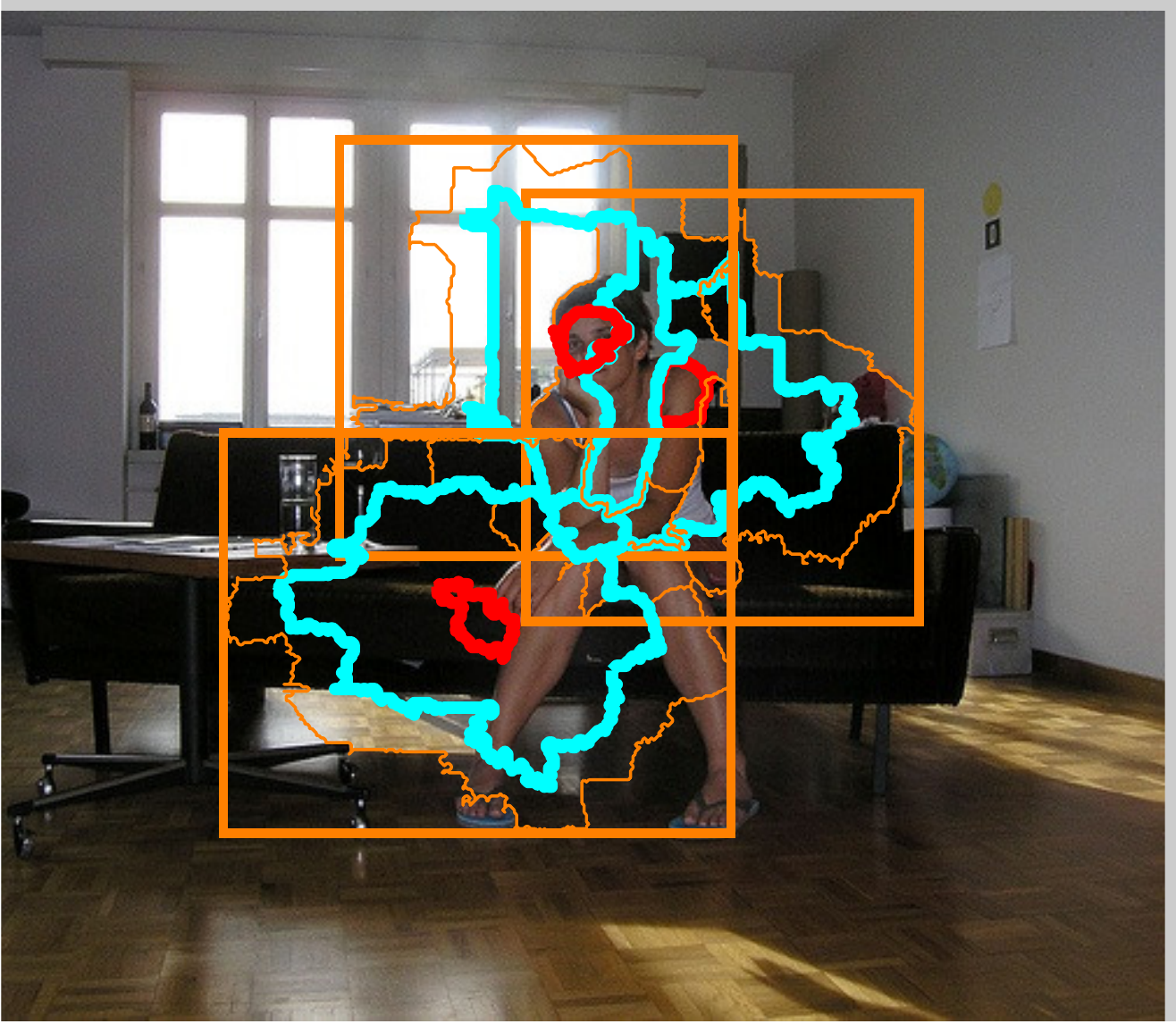}
  \end{tabular}

  \caption{Examples of zoom-out regions: red for
    \textcolor{red}{superpixel}, cyan for \textcolor{cyan}{proximal}
    region, solid orange for \textcolor{orange}{distant} region;
    curved orange line shows the extent of the radius-3 neighborhood
    on which the distant region is based. The
    \textcolor{black!20!green}{global} region is always the entire
    image. The image on the left shows superpixel boundaries in black;
  these are typical. Distant regions tend to enclose large
  portions of objects. Proximal regions are more likely to include moderate
  portions of objects, and both often include surrounding
  objects/background as well.}
  \label{fig:regions}
\end{figure*}

We cast category-level segmentation of an image as classifying a set
of superpixels. Since we expect to apply the same classification
machine to every superpixel, we would like the nature of the superpixels
to be similar, in particular their size. In our experiments we use
SLIC~\cite{achanta_pami12}, but other methods that produce
nearly-uniform grid of superpixels might work similarly
well. Figures~\ref{fig:regions} and~\ref{fig:region-examples} provide a few illustrative examples
for this discussion.

\subsection{Scoping the zoom-out features}

The main idea of our zoom-out architecture is to allow features
extracted from different levels of spatial context around the
superpixel to contribute to labeling decision at that superpixel. To
this end we define four levels of spatial extent. For each of these we
discuss the role we intend it to play in the eventual classification
process, and in particular focus on the expected relationships between the features
at each level computed for different superpixels in a given image. We
also briefly comment on the kind of features that we may want to
compute at each level.

\subsubsection{Local zoom (close-up)} The narrowest scope is the superpixel itself. We
expect the features extracted here to capture local evidence for or
against a particular labeling: color, texture, presence of small
intensity/gradient patterns, and other properties computed over a
relatively small contiguous set of pixels, could all contribute at this level. The
local features may be quite different even for neighboring
superpixels, especially if these straddle category or object boundaries.

\subsubsection{Proximal zoom} The next scope is a region that surrounds the
superpixel, perhaps an order of magnitude larger in area. We expect
similar kinds of visual properties to be informative as with the local features,
but computed over larger proximal regions, these may have different statistics, for instance, we can expect various histograms (e.g., color) to be
less sparse. Proximal features may capture
information not available in the local scope; e.g., for locations at the
boundaries of objects they will represent the appearance of both
categories. For classes with non-uniform appearance they may better
capture characteristic distributions for that class. This scope is
usually 
still too myopic to allow us to reason confidently about presence of objects.

Two neighboring
superpixels could still have quite different proximal features,
however some degree of smoothness is likely to arise from the
significant overlap between neigbhors' proximal regions.
As an example,
consider color features over the body of a leopard; superpixels for individual dark brown spots might
appear quite different from their neighbors (yellow fur) but their
proximal regions will have pretty similar distributions (mix of yellow
and brown).
Superpixels that are sufficiently far from each other could still, of
course, have drastically different proximal features. 

\input{fig2}

\subsubsection{Distant zoom} Zooming out further, we move to the distant level:
a region large enough to include sizeable fractions of objects, and
sometimes entire objects. At this level our scope is wide enough to
allow reasoning about shape, presence of more complex patterns in color and
gradient, and the spatial layout of such patterns. Therefore we can expect
more complex features that represent these properties to be useful here.
Distant regions are more
likely to straddle true boundaries in the image, and so this
higher-level feature extraction may include a significant area in both
the category of the superpixel at hand and nearby categories. For
example, consider a person sitting on a chair; bottle on a dining
table; pasture animals on the background of grass, etc. Naturally we
expect this to provide useful information on both the appearance of a
class and its context.

For neighboring superpixels, distant regions will have a very large
overlap; superpixels which are one superpixel apart will have
significant, but lesser, overlap in their distant regions; etc., all
the way to superpixels that are sufficiently far apart that the
distant regions do not overlap and thus are independent given the scene. This
overlap in the regions is likely to lead to somewhat gradual changes
in features, and to impose an entire network of implicit smoothness
``terms'', which depend both on the distance in the image and on the
similarity in appearance in and around superpixels. Imposing
such smoothness in a CRF usually leads to
a very complex, intractable model.

\subsubsection{Global zoom} The final zoom-out scope is the entire scene. Features computed at this level capture ``what
kind of an image'' we are looking at. One aspect of this global
context is image-level classification: since state of the art in image
classification seems to be dramatically higher than that of detection
or segmentation~\cite{everingham2014pascal,russakovsky2014imagenet}
 we can expect
image-level features to help determine presence of categories in the
scene and thus guide the segmentation.

More subtly, features that are useful for classification can be
directly useful for global support of local labeling decisions; e.g.,
lots of green in an image supports labeling a (non-green) superpixel
as cow or sheep more than it supports labeling that superpixel as
chair or bottle, other things being equal. On the other hand, lots of
straight vertical lines in an image would perhaps suggest man-made
environment, thus supporting categories relevant to indoors or urban
scenes more than, say, wildlife.

At this global level, all superpixels in an image will of course have the same
features, imposing (implicit, soft) global constraints. This is yet another form
of high-order interaction that is hard to capture in
a CRF framework, despite numerous attempts~\cite{boix_ijcv12}.

\subsection{Learning to label with asymmetric loss}
Once we have computed the zoom-out features we simply concatenate them
into a feature vector representing a superpixel. For superpixel $s$ in
image $\mathcal{I}$, we will denote this feature vector as
\begin{equation}
\bfgreek{phi}_{\text{zoom-out}}(s,\mathcal{I})\,=\,
\begin{bmatrix}
\bfgreekco{phi}{red}_{\textcolor{red}{loc}}(s,\mathcal{I})\\
\bfgreekco{phi}{cyan}_{\textcolor{cyan}{prox}}(s,\mathcal{I})\\
\bfgreekco{phi}{orange}_{\textcolor{orange}{dist}}(s,\mathcal{I})\\
\bfgreekco{phi}{black!20!green}_{\textcolor{black!20!green}{glob}}(\mathcal{I})
\end{bmatrix}
\end{equation}
For the training data, we will associate a single category label $y_s$
with
each superpixel $s$. This decision carries some risk, since in any non-trivial
over-segmentation some of the superpixels will not be perfectly
aligned with ground truth boundaries. In section~\ref{sec:exp} we
evaluate this risk empirically for our choice of superpixel settings
and confirm that it is indeed minimal.

Now we are ready to train a classifier that maps $s$ in image
$\mathcal{I}$ to $y_s$ based on $\bfgreek{phi}_{\text{zoom-out}}$;
this requires choosing the empirical loss function to be minimized,
subject to regularization.
In semantic
segmentation settings, a factor that must impact this choice is the highly imbalanced nature of the labels. Some categories are
much more common than others, but our goal (encouraged by the
way benchmark like VOC evaluate segmentations) is to predict them
equally well. It is well known that training on imbalanced data
without taking precautions can lead to
poor
results~\cite{farabet2013learning,pinheiro2014recurrent,lempitsky2011pylon}. A
common way to deal with this is to stratify the
training data; in practice this means that we throw away a large
fraction of the data corresponding to the more common
classes. We follow an alternative which we find less wasteful, and
which in our experience often produces dramatically better results:
use all the data, but change the loss. There has been some work on
loss design for learning segmentation~\cite{tarlow2012structured}, but
the simple weighted loss we describe below has to our knowledge been missed in
segmentation literature, with the exception
of~\cite{kuettel2012segmentation} and~\cite{lempitsky2011pylon}, where
it was used for binary segmentation.

Let the frequency of class $c$ in the
training data be
$f_c$, with $\sum_cf_c=1$. Suppose our choice of loss is log-loss; we
modify it to be
\begin{equation}
  \label{eq:asymloss}
  -\frac{1}{N}\sum_{i=1}^N\frac{1}{f_{y_i}}\log \widehat{p}\left(y_i|\bfgreek{phi}(s_i,\mathcal{I}_i)\right),
\end{equation}
where $\widehat{p}\left(y_i|\bfgreek{phi}(s_i,\mathcal{I}_i)\right)$
is the estimated probability of the correct label for segment $s_i$ in
image $\mathcal{I}_i$, according to our model. In other words, we scale the loss by the inverse frequency of
each class, effectively giving each pixel of less frequent classes
more importance. This modification does not change loss convexity, and
only requires minor changes in the optimization code, e.g., back-propagation.

%% file: fig2.tex
\begin{figure*}[!tb]
\begin{minipage}{.2\textwidth}
\begin{tabular}{>{\hspace{-2em}}c>{\hspace{-2em}}c>{\hspace{-2em}}c>{\hspace{-2em}}c}
    \begin{tabular}{c}
        \includegraphics[width=\textwidth]{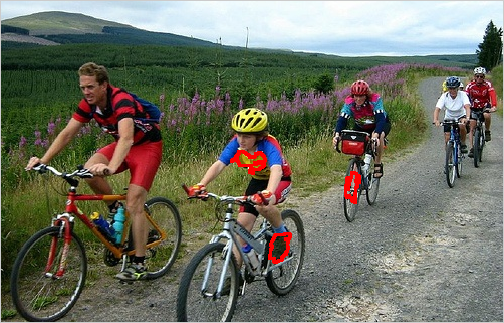}\\
    \begin{tabular}{ccc}
        \includegraphics[width=.3\textwidth]{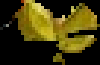} &
        \includegraphics[width=.3\textwidth]{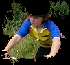} &
        \includegraphics[width=.3\textwidth]{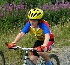} \\
        \includegraphics[width=.24\textwidth]{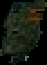} &
        \includegraphics[width=.3\textwidth]{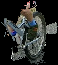} &
        \includegraphics[width=.3\textwidth]{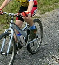} \\
        \includegraphics[width=.18\textwidth]{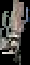} &
        \includegraphics[width=.3\textwidth]{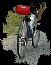} &
        \includegraphics[width=.3\textwidth]{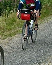}
    \end{tabular}
    \end{tabular} &
    \begin{tabular}{c}
    \includegraphics[width=.78\textwidth]{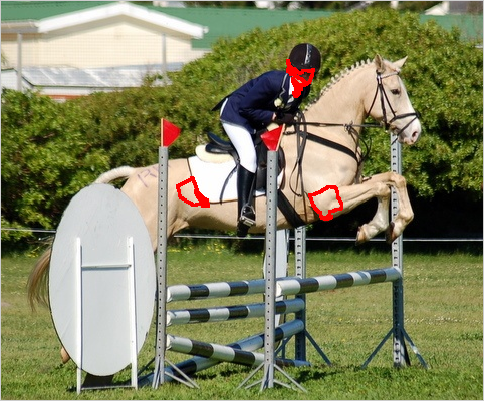}\\
    \begin{tabular}{ccc}
        \includegraphics[width=.3\textwidth]{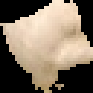} &
        \includegraphics[width=.3\textwidth]{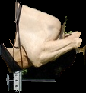} &
        \includegraphics[width=.3\textwidth]{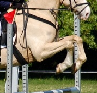} \\
        \includegraphics[width=.24\textwidth]{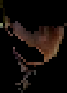} &
        \includegraphics[width=.3\textwidth]{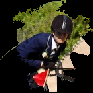} &
        \includegraphics[width=.3\textwidth]{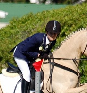} \\
        \includegraphics[width=.3\textwidth]{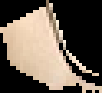} &
        \includegraphics[width=.3\textwidth]{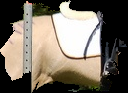} &
        \includegraphics[width=.3\textwidth]{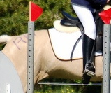}
    \end{tabular}
    \end{tabular} &
    \begin{tabular}{c}
    \includegraphics[width=.78\textwidth]{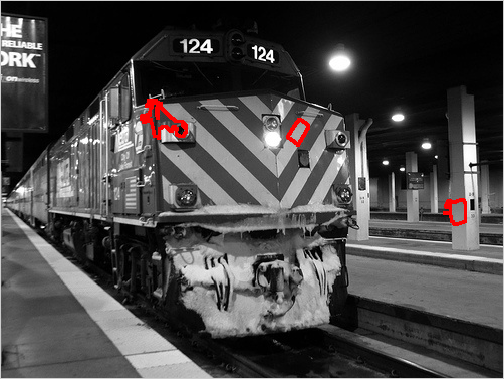}\\
    \begin{tabular}{ccc}
        \includegraphics[width=.3\textwidth]{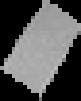} &
        \includegraphics[width=.3\textwidth]{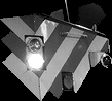} &
        \includegraphics[width=.3\textwidth]{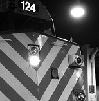} \\
        \includegraphics[width=.24\textwidth]{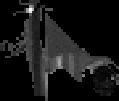} &
        \includegraphics[width=.3\textwidth]{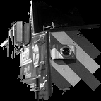} &
        \includegraphics[width=.3\textwidth]{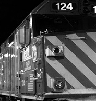} \\
        \includegraphics[width=.3\textwidth]{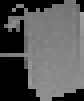} &
        \includegraphics[width=.3\textwidth]{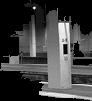} &
        \includegraphics[width=.3\textwidth]{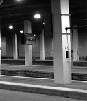}
    \end{tabular}
    \end{tabular} &
\begin{tabular}{c}
    \includegraphics[width=.78\textwidth]{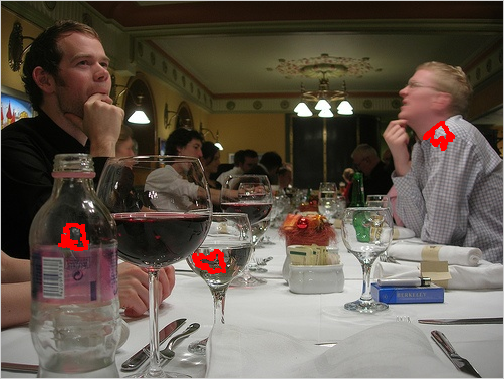}\\
    \begin{tabular}{ccc}
        \includegraphics[width=.3\textwidth]{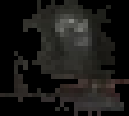} &
        \includegraphics[width=.3\textwidth]{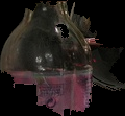} &
        \includegraphics[width=.3\textwidth]{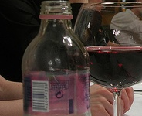} \\
        \includegraphics[width=.24\textwidth]{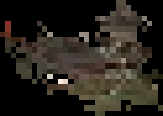} &
        \includegraphics[width=.3\textwidth]{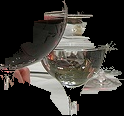} &
        \includegraphics[width=.3\textwidth]{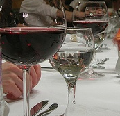} \\
        \includegraphics[width=.3\textwidth]{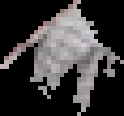} &
        \includegraphics[width=.3\textwidth]{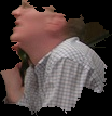} &
        \includegraphics[width=.3\textwidth]{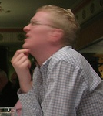}
    \end{tabular}
    \end{tabular} 
\end{tabular}
\end{minipage}
    \caption{Showing three superpixels in each image (top), followed
      by corresponding zoom-out regions that are seen by the
      segmenation process, (left) superpixel, (center) proximal
      region, (right) distant. As we zoom in from image to
      the superpixel level, it becomes increasingly hard to tell what
      we are looking at, however the higher zoom-out levels provide
      rich contextual information.}
 \label{fig:region-examples}
\end{figure*}

%% file: related.tex
\section{Related work}\label{sec:related}

The literature on segmentation is vast, and here we only mention work
that is either significant as having achieved state of the art
performance in recent times, or is closely related to ours in some
way. In Section~\ref{sec:exp} we compare our performance to
that of most of the methods mentioned here.

Many prominent segmentation methods rely on conditional random fields (CRF) over nodes
corresponding to pixels or superpixels.
Such models incorporate local evidence in unary potentials, while
interactions between label assignments are captured by pairwise and
possibly higher-order potentials. This includes various hierarchical CRFs~\cite{shotton_ijcv09,ladicky_iccv09,lempitsky2011pylon,boix_ijcv12}. In
contrast, we let the zoom-out features (in CRF terminology, the unary potentials) to capture higher-order structure.

Another recent trend has been to follow a multi-stage
approach: First a set of proposal regions is generated, by a
category-independent~\cite{carreira2012cpmc,UijlingsIJCV2013} or
category-aware~\cite{arbelaez_cvpr12} mechanism.
Then the regions are scored or ranked based on their compatibility
with the target classes. Work in
this vein
includes~\cite{carreira2012semantic,ion2011probabilistic,arbelaez_cvpr12,carreira_ijcv12,Li2013codemaps}.
A similar approach is taken
in~\cite{yadollahpour2013discriminative}, where multiple segmentations
obtained from~\cite{carreira2012semantic}
are re-ranked using a discriminatively trained model.
Recent advances along these lines include~\cite{hariharan2014simultaneous}, which
uses convnets
and~\cite{dhruv-new}, which improves upon the re-ranking
in~\cite{yadollahpour2013discriminative}, also using
convnet-based features. At submission time, these two lines of work are
roughly tied for the previous state of the art\footnote{before this work and
concurrent efforts, all relying on multi-level features},
with mean accuracy of 51.6\% and 52.2\%, respectively, on VOC 2012 test.
In contrast to most of the work in this group, we do not rely on
region generators, and limit preprocessing to
over-segmentation of the image into a large number of superpixels. 

A body of work examined the role of
context and the importance of non-local evidence in segmentation. The
idea of computing features over a neighborhood of superpixels for
segmentation purposes was
introduced in~\cite{fulkerson2009class} and~\cite{lim2009context};
other early work on using forms of context for segmentation
includes~\cite{shotton_ijcv09}. 
A study in~\cite{lucchi2011spatial} concluded that non-unary terms may be unnecessary when neighborhood and global
information is captured by unary terms, but the results were
significantly inferior to state of the art at the time. 

Recent work closest to ours
includes~\cite{farabet2013learning,pinheiro2014recurrent,socher2011parsing,mostajabi2014robust,huang2013deep}. 
In~\cite{farabet2013learning}, the same convnet is applied on
different resolutions of the image and
combined with a tree-structured graph over superpixels to impose
smoothness. In~\cite{mostajabi2014robust} the features applied to
multiple levels (roughly analogous to our local+proximal+global) are
also same, and hand-crafted instead of using convnets. In~\cite{pinheiro2014recurrent} there is also a single
convnet, but it is applied in a recurrent fashion, i.e., input to the
network includes, in addition to the scaled image, the feature maps
computed by the network at a previous level. A similar idea is pursued
in~\cite{huang2013deep}, where it is applied to boundary detection in
3D biological data. In contrast with all of these, we use different
feature extractors across levels, some of them with a much smaller
receptive field than any of the networks in the
literature. We show in
Section~\ref{sec:exp} that our approach obtains better performance (on
Stanford Background Dataset) than that reported
for~\cite{farabet2013learning,pinheiro2014recurrent,socher2011parsing,mostajabi2014robust};
no comparison to~\cite{huang2013deep} is available.

Finally, two pieces of concurrent work share some key ideas with our
work, and we discuss these in some detail
here. In~\cite{long2014fully}, achieving 62.2\% mean accuracy on VOC
2012 test, a 16-layer convnet is applied to
an image at a coarse grid of locations. Predictions
based on the final layer of the network are upsampled and summed with predictions made from
intermediate layers of the network. Since units in lower layers are associated with
smaller receptive field than those in the final layer, this mechanism
provides fusion of information across spatial levels, much like our
zoom-out features. The architecture in~\cite{long2014fully} is more
efficient than our current implementation since it reuses computation
efficiently (via framing the computation of features at multiple
locations as convolution), and since the entire model is trained end-to-end.
However, it relies in a somewhat reduced range of spatial context
compared to our work; using our terminology proposed above, the architecture in~\cite{long2014fully}
roughly analogous to combining our
proximal and distant zoom-out levels, along with another one above
distant, but without the global level\footnote{assuming the original
  image is larger than the net receptive field, which is typically the
  case. In contrast, we zoom out far enough to include the entire
  image, which we resize to the desired receptive field size.}, important to establish the
general scene context, and without the local level, important for precise
localization of boundaries. Also, we choose to ``fuse'' not the
predictions made at different levels, but the rich evidence (features
themselves).  

The other recent work with significant similarities to ours
is~\cite{hariharan2014hypercolumns}, where \emph{hypercolumns} are
formed by pooling evidence extracted from nested regions around a
pixel; these too resemble our zoom-out feature
representations. In contrast with~\cite{long2014fully} and with our
work, the input to the system here consists of a hypothesized
detection bounding box, and not an entire image. The hypercolumn
includes some local information (obtained from the {\tt pool2} layer)
as well as information pooled from regions akin to our proximal as
well as the level that is global relative to the hypothesized bounding
box, but is not global with respect to the entire image. A further
difference from our work is the use of location-specific classifiers
within the bounding box, instead
of the same classifer applied everywhere. This work achieves 59.0\% mean
accuracy on VOC 2012 test.

%% file: experiments.tex
\section{Experiments}\label{sec:exp}

Our main set of experiments focuses on the PASCAL VOC category-level segmentation
benchmark with 21 categories, including the catch-all background
category. VOC is widely considered to be the main semantic
segmentation benchmark today\footnote{The Microsoft Common Objects in
  Context (COCO) promises to become another such benchmark, however at
  the time of this writing it is not yet fully set up with test set
  and evaluation procedure}. The original data set labeled with segmentation ground truth
consists of {\tt train}
and {\tt val} portions
(about 1,500 images in each). Ground
truth labels for additional 9,118 images have been provided by authors
of~\cite{BharathICCV2011}, and are commonly used in training segmentation models.
In all experiments below, we used the combination of these additional images with the original
{\tt train}
set for training, and {\tt val} was used only as held out validation set, to tune
parameters and to perform ``ablation studies''.

The
main measure of success is accuracy on the test, which for VOC 2012
consists of 1,456 images. No ground truth is available for test, and
accuracy on it can only be obtained by uploading predicted
segmentations to the evaluation server.
The standard evaluation measure for category-level segmentation in VOC
benchmarks is per-pixel accuracy, defined as intersection of the
predicted and true sets of pixels for a given class, divided by their
union (IU in short). This is averaged across the 21 classes to provide a single accuracy
number, mean IU, usually used to measure overall performance of a method.

\subsection{Superpixels and neighborhoods}
We obtained roughly 500 SLIC superpixels~\cite{achanta_pami12} per
image (the exact number varies per image), with
the parameter $m$ that controls the tradeoff between spatial and color
proximity set to 15, producing superpixels which tend to be of uniform
size and regular shape, but adhere to local
boundaries when color evidence compels it. This results
in average supepixel region of 21$\times$21 pixels. An example of a
typical over-segmentation is shown in Figure~\ref{fig:regions}.

Proximal region for a superpixel $s$ is defined as a set of
superpixels within radius 2 from $s$, that is, the immediate neighbors
of $s$ as well as their immediate neighbors. The proximal region can
be of arbitrary shape; its average size in training images is 100x100 pixels.
The distant region is
defined by all neighbors of $s$ up to the 3rd degree, and consists of
the bounding box around those neighbors, so that in contrast to
proximal, it is always rectangular; its average size is 170$\times$170
pixels.
Figure~\ref{fig:regions} contains a
few typical examples for the regions.

\subsection{Zoom-out feature computation}
Feature extraction differs according to the zoom-out level, as
described below.

\subsubsection{Local features}

To represent a superpixel we use a number of well known features as well
as a small set of learned features.

\begin{description}
\item[Color]
We compute histograms separately for each of the three L*a*b color
channels, using 32 as well as 8 bins, using equally spaced binning;
this yields 120 feature dimensions.  We also compute the entropy of
each 32-bin histogram as an additional scalar feature (+3 dimensions). Finally, we
also re-compute histograms using adaptive binning, based on observed quantiles
in each image (+120 dimensions).

\item[Texture]
Texture is represented by histogram of texton assignments, with
64-texton dictionary computed over a sampling of images from the
training set. This histogram is augmented with the value of its
entropy. In total there are 65 texture-related channels.

\item[SIFT] A richer set of features capturing appearance
is based on ``bag of words'' representations computed over SIFT
descriptors. The descriptors are computed over a regular grid (every 8
pixels), on 8- and 18-pixel patches, separately for each L*a*b
channel. All the descriptors are assigned to a dictionary of 500
visual words. Resulting assignment histograms are averaged for two
patch sizes in each channel, yielding a total of 1500 values, plus 6
values for the entropies of the histograms.

\item[Location] Finally, we encode superpixel's location, by
computing its image-normalized coordinates relative to the center as
well as its shift from the center (the absolute value of the
coordinates); this produces four feature values.

\item[Local convnet] Instead of using hand-crafted features we
could learn a representation for the superpixel using a convnet. We
trained a network with 3 convolutional (conv) + pooling + RELU layers,
with 32, 32 and 64 filters respectively, followed by two
fully connected layers (1152 units each) and finally a softmax layer. The input to this
network is the bounding box of the superpixel, resized to 25$\times$25
pixels and padded to $35\times35$, in L*a*b color space. The filter
sizes in all layers are 5$\times$5; the pooling layers all have 3$\times$3
receptive fields, with stride of 2. We trained the network using
back-propagation, with the objective of minimizing log-loss for
superpixel classification. The output of the softmax layer of this
network is used as a 21-dimensional feature vector. Another
network with the same architecture was trained on binary
classification of foreground vs. background classes; that gives us two
more features.
\end{description}


\subsubsection{Proximal features}
We use the same set of handcrafted features as for local regions,
resulting in 1818 feature dimensions.

\subsubsection{Distant and global features}
For distant and global features we use deep convnets originally
trained to classify images. In our initial experiments we used the CNN-S network
in~\cite{simonyan2014return}. It has 5 convolution layers
(three of them followed by pooling) and two fully connected layers. To extract the relevant features, we
resize either the distant region or the entire image to
224$\times$224 pixels, feed it to the network, and record the
activation values of the last fully-connected layer with 4096 units.
In a subsequent set of experiments we switched from CNN-S to a 16
layer network introduced in~\cite{simonyan2014very}. This network,
which we refer to as VGG-16, contains more layers that apply
non-linear transformations, but with smaller filters, and thus may learn richer representation
with fewer parameters. It has
produced excellent results on image classification and recently has
been reported to lead to a much better performance when used in
detection~\cite{girshick2014rich}. As reported below, we also observe
a significant improvement when using VGG-16 to extract distant
and global features, compared to performance with CNN-S.
    Note that both networks we used were originally trained for 1000-category ImageNet
classification task, and we did not fine-tune it in any way on VOC data.

\subsection{Learning setup}
With more than 10,000 images and roughly 500 superpixels per image, we
have more than 5 million training examples. 

In section~\ref{sec:method} we mentioned an obvious concern when reducing image labeling problem to superpixel
labeling is whether this leads to loss of achievable accuracy, since
superpixels need not perfectly adhere to true boundaries. Having
assigned each superpixel a category label based on the majority of
pixels in it, we computed the accuracy of this assignment on val: 94.4\%. Since accuracies of most of today's methods are well below this
number, we can assume that any potential loss of accuracy due to our
commitment to superpixel boundaries is going to play a minimal role
in our results.

We trained the classifiers mentioned below with asymmetric
loss~\eqref{eq:asymloss} using
Caffe~\cite{jia2014caffe}, on a single machine equipped with a Tesla K40
GPU. During training we used fixed learning rate of 0.0001, and
weight decay factor of 0.001.


\subsection{Results on PASCAL VOC 2012}
To empirically assess the importance of features extracted
at different zoom-out levels, we trained linear (softmax) models
using various feature
subsets, as shown in Table~\ref{tab:ablation}, and evaluated them on
VOC 2012 {\tt val}. In these experiments we used CNN-S to extract
distant and global features.

\input{colorcode}

\begin{table}[!th]
  \centering
{\small
  \begin{tabular}{|l|l|}
    \hline
Feature set & mean accuracy\\
\hline
local &  14.6\\
proximal & 15.5\\
local+proximal & 17.7\\
local+distant & 37.38\\
local+global & 41.8\\
local+proximal+global & 43.4\\
distant+global & 47.0\\
\hline
full zoom-out, symmetric loss & 20.4\\
\hline
full zoom-out, asymmetric loss & {\bf 52.4}\\
\hline
  \end{tabular}}
  \caption{Ablation study: importance of features from different
    levels under linear superpixel classification. Results on VOC 2012 val, mean class accuracy.}
  \label{tab:ablation}
\end{table}

\begin{figure*}
  \centering
  \begin{tabular}{cc>{\hspace{-1.5em}}c>{\hspace{-2.5em}}c>{\hspace{-2.5em}}c>{\hspace{-2.5em}}c}
    \raisebox{.8em}{\includegraphics[width=.15\textwidth]{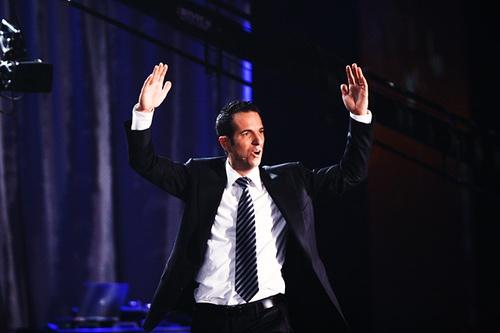}} &
    \raisebox{.8em}{\includegraphics[width=.15\textwidth]{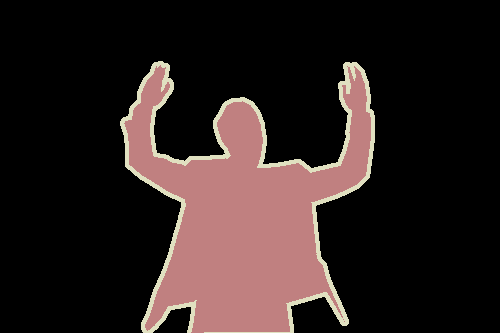}} &
    \includegraphics[width=.2\textwidth]{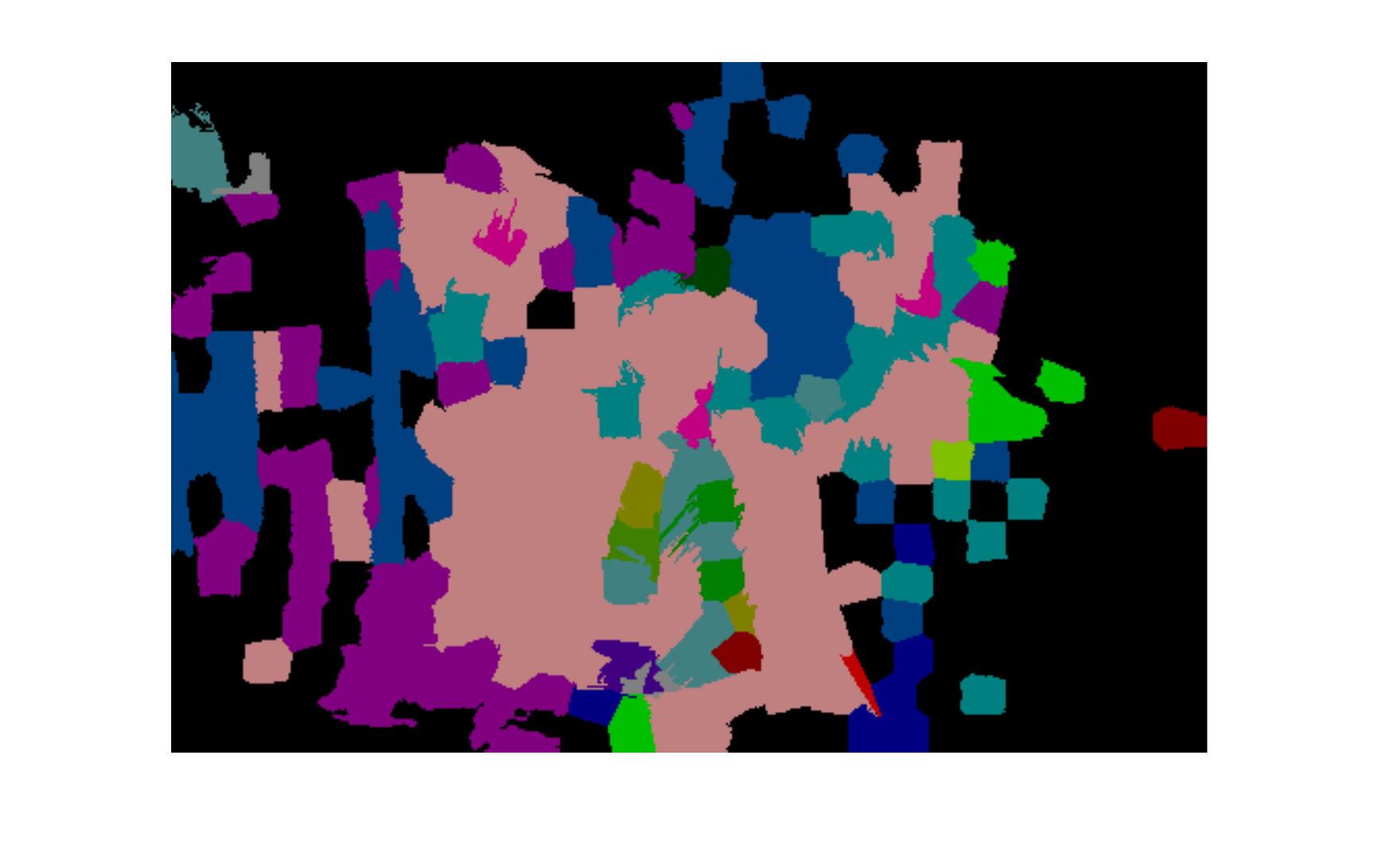} &
    \includegraphics[width=.2\textwidth]{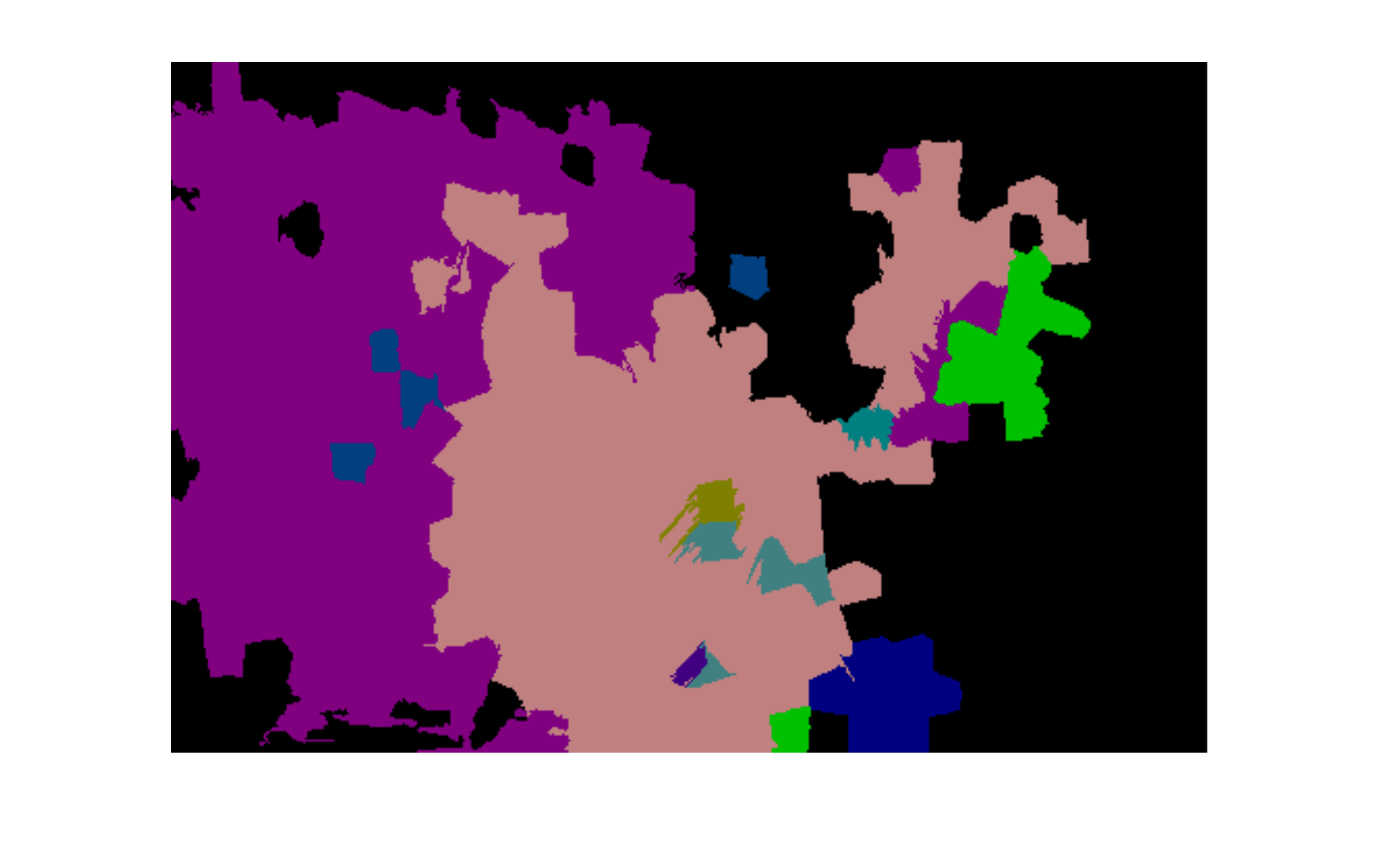} &
    \includegraphics[width=.2\textwidth]{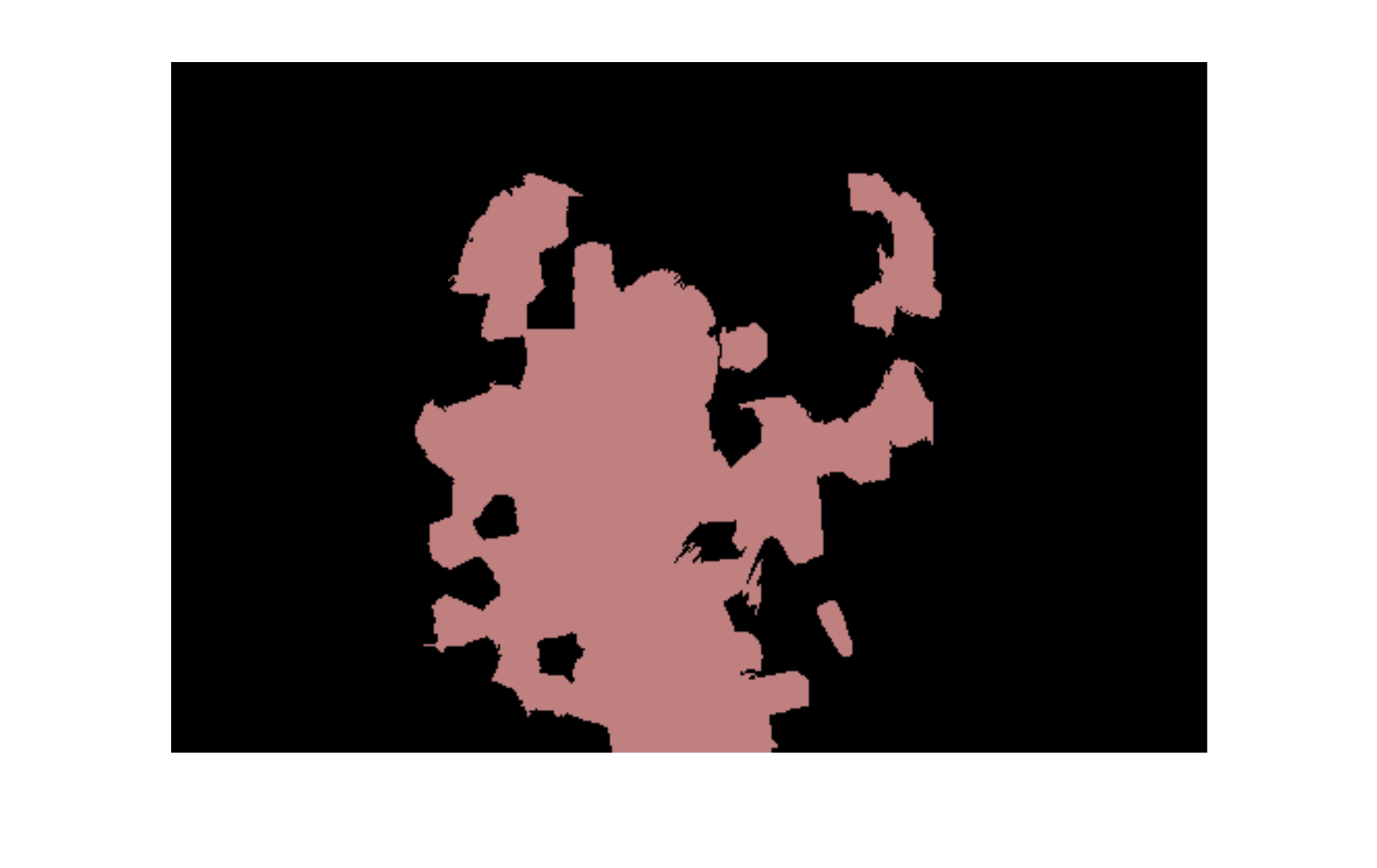} &
    \includegraphics[width=.2\textwidth]{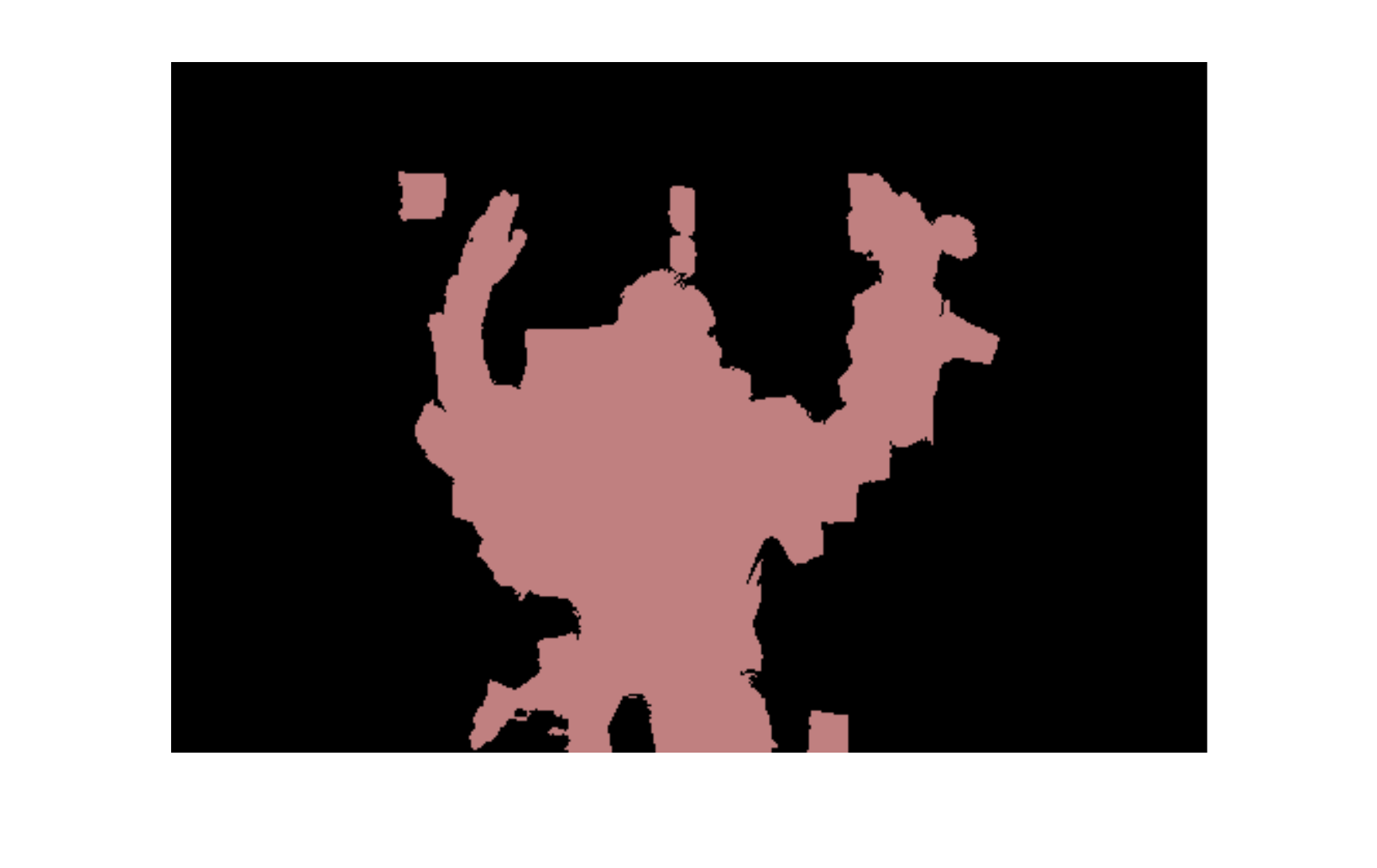}\\
    \raisebox{.8em}{\includegraphics[width=.15\textwidth]{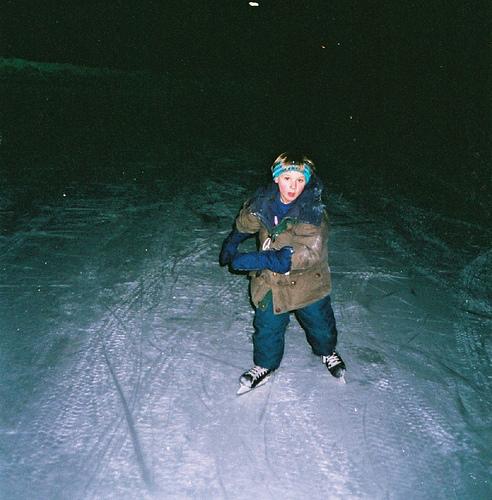}} &
    \raisebox{.8em}{\includegraphics[width=.15\textwidth]{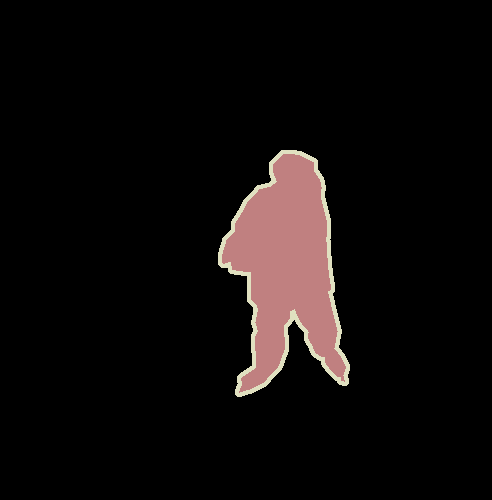}} &
    \includegraphics[width=.2\textwidth]{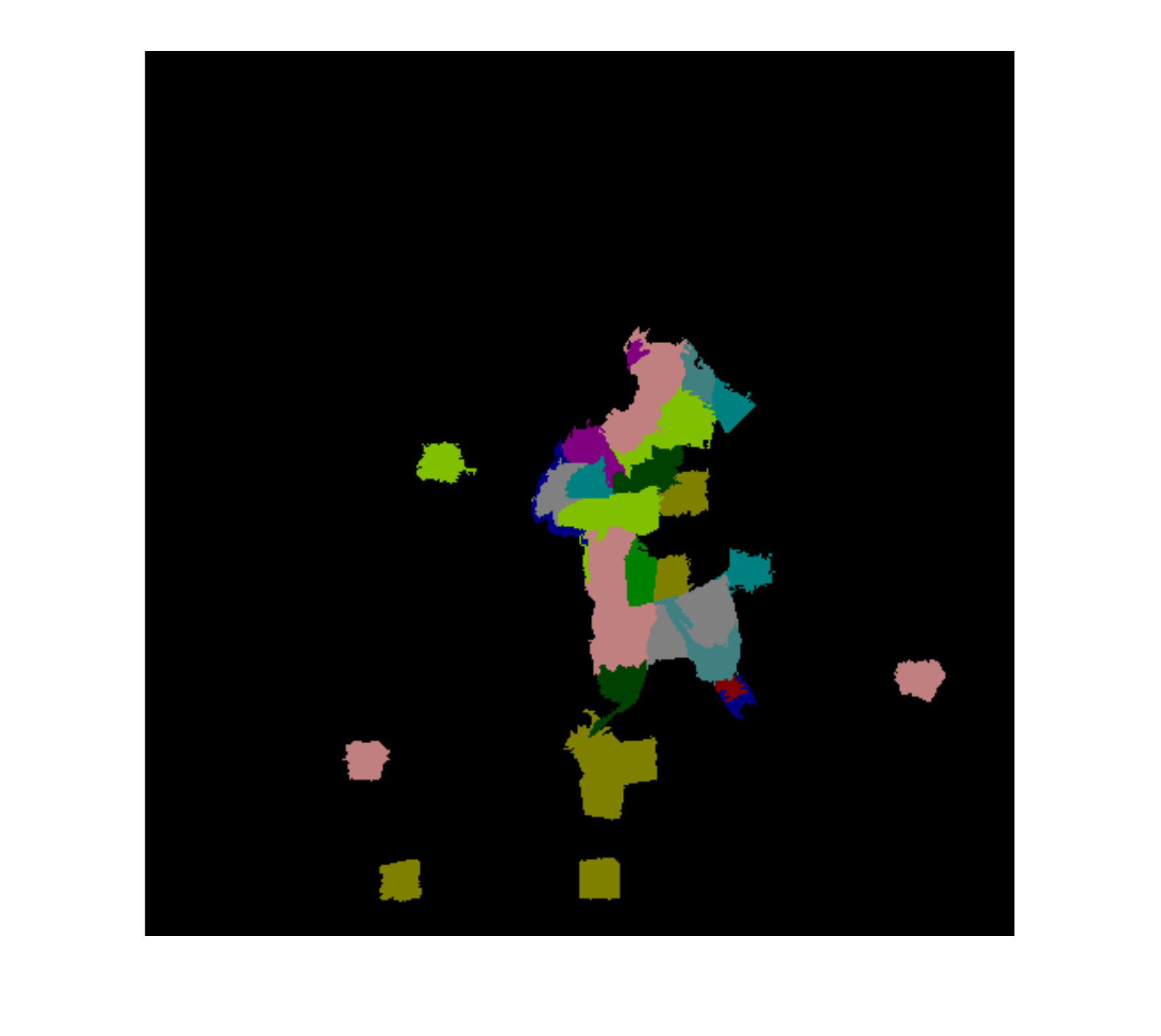} &
    \includegraphics[width=.2\textwidth]{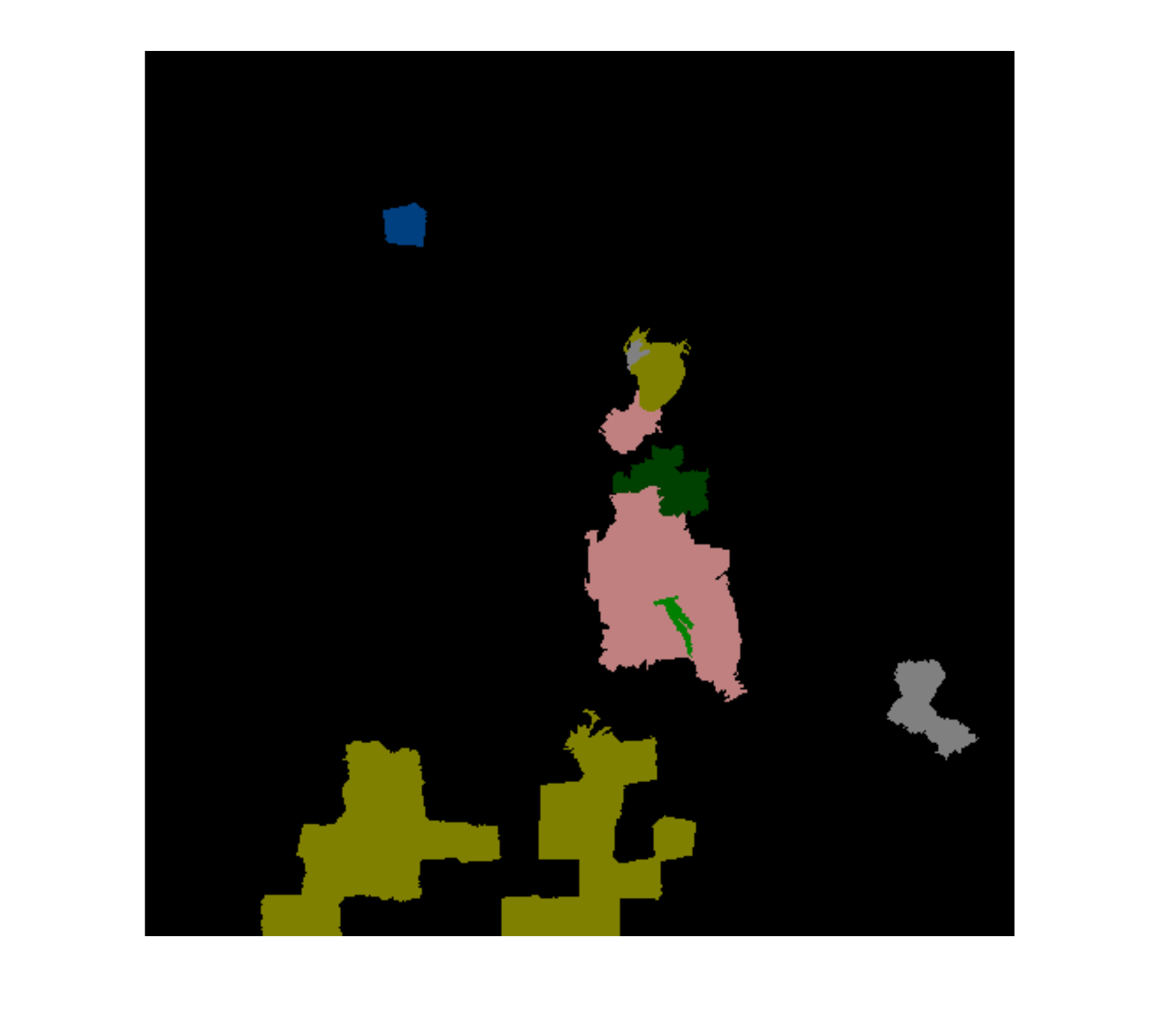} &
    \includegraphics[width=.2\textwidth]{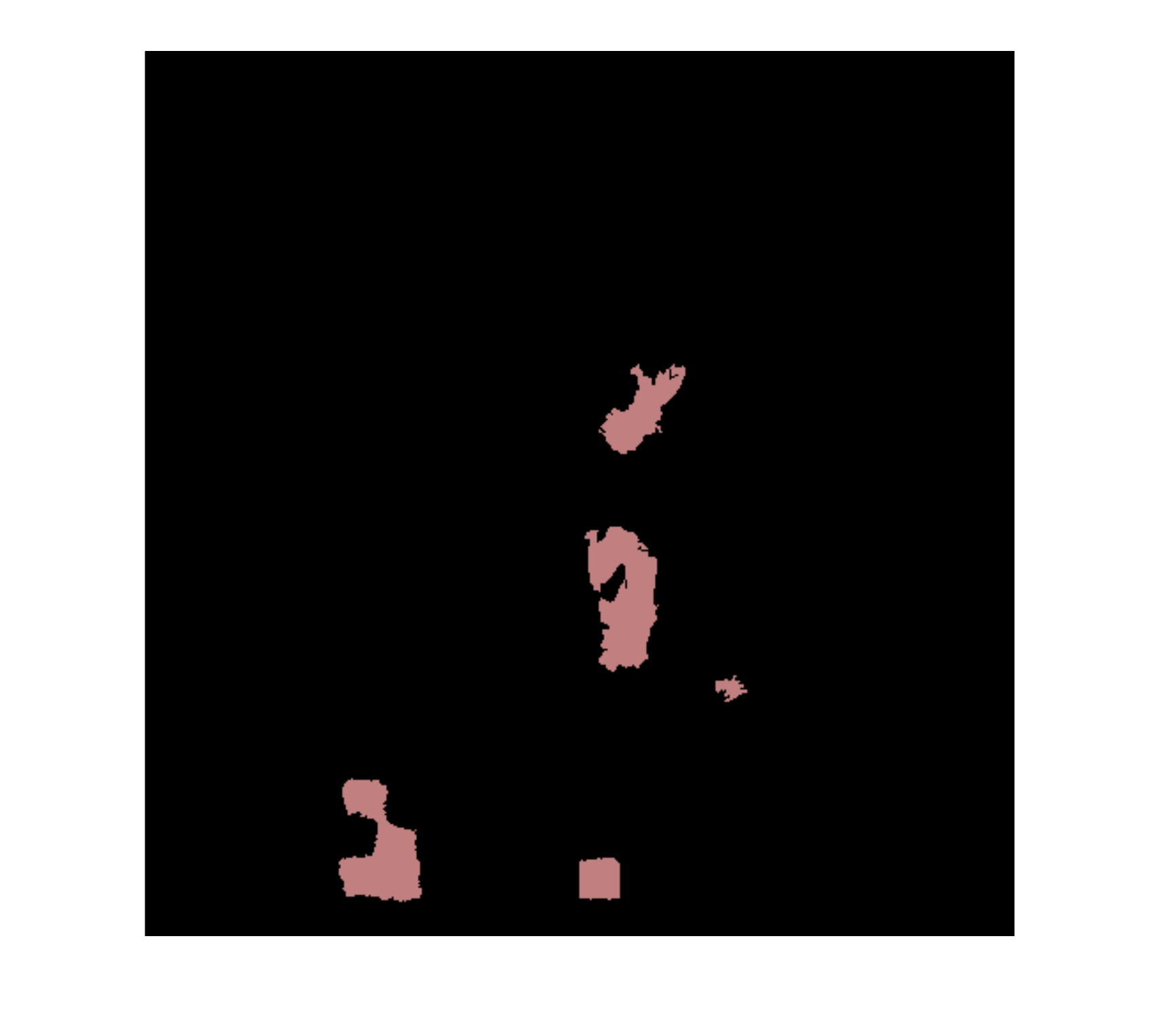} &
    \includegraphics[width=.2\textwidth]{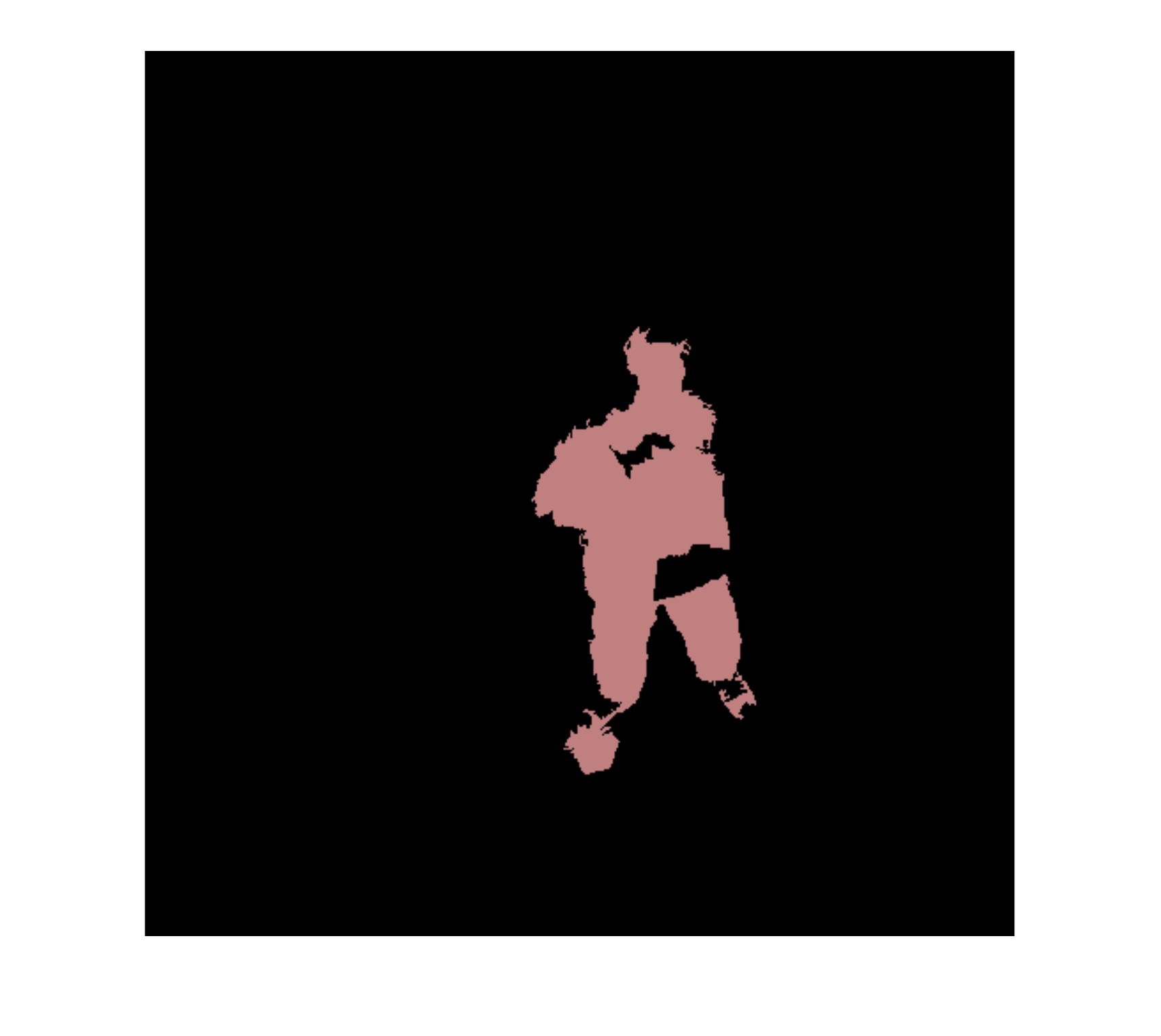}\\
    \raisebox{.8em}{\includegraphics[width=.15\textwidth]{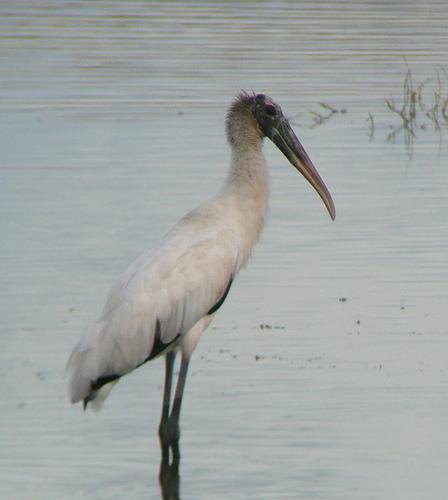}} &
    \raisebox{.8em}{\includegraphics[width=.15\textwidth]{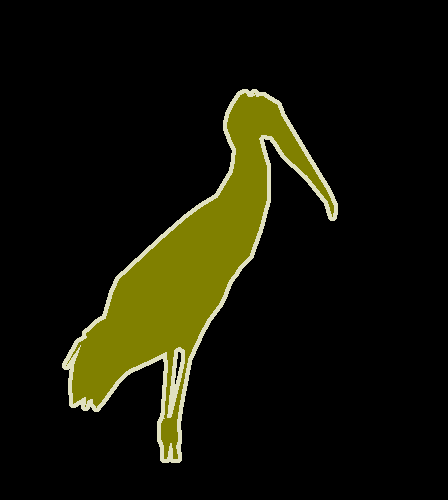}} &
    \includegraphics[width=.2\textwidth]{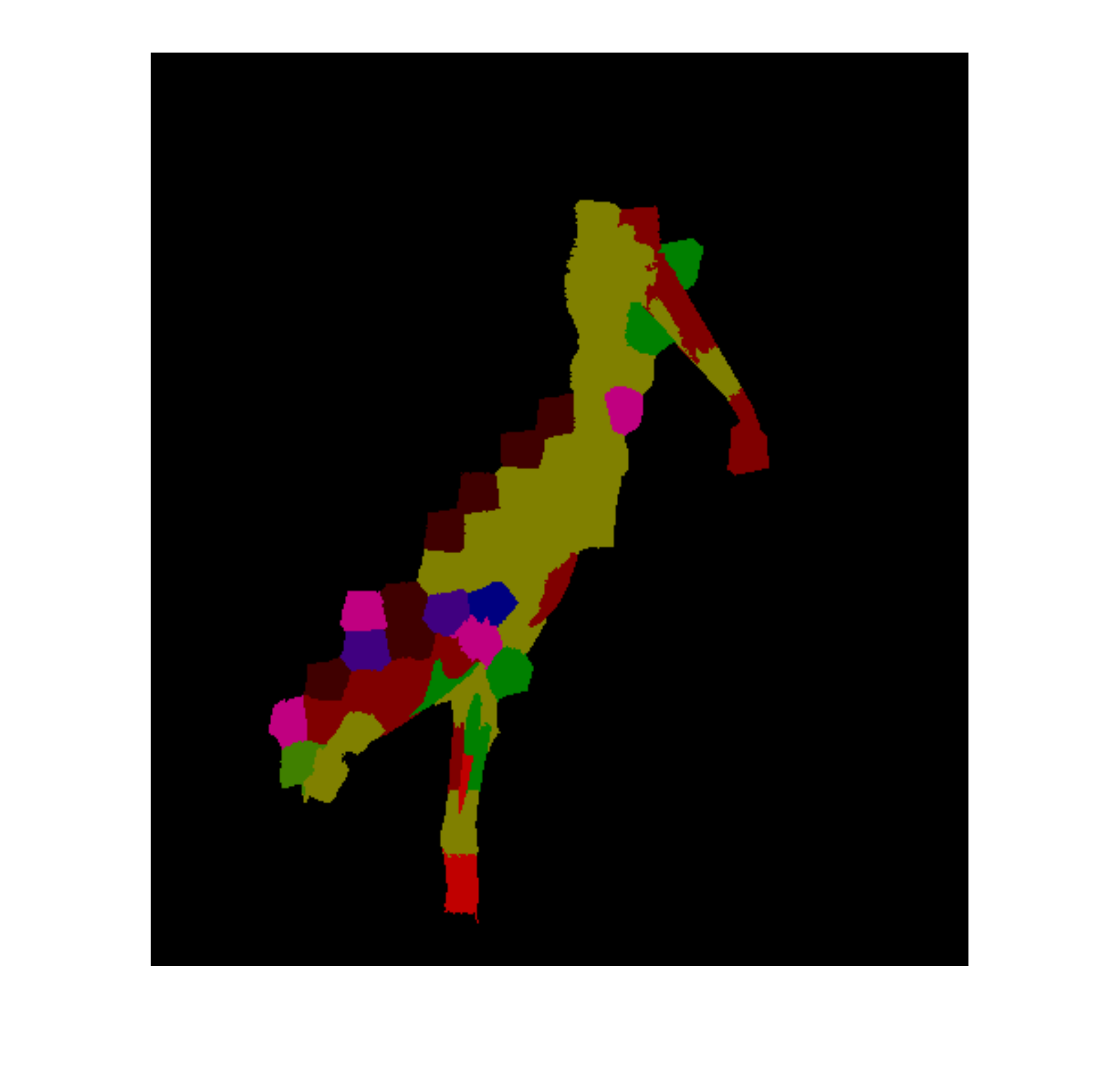} &
    \includegraphics[width=.2\textwidth]{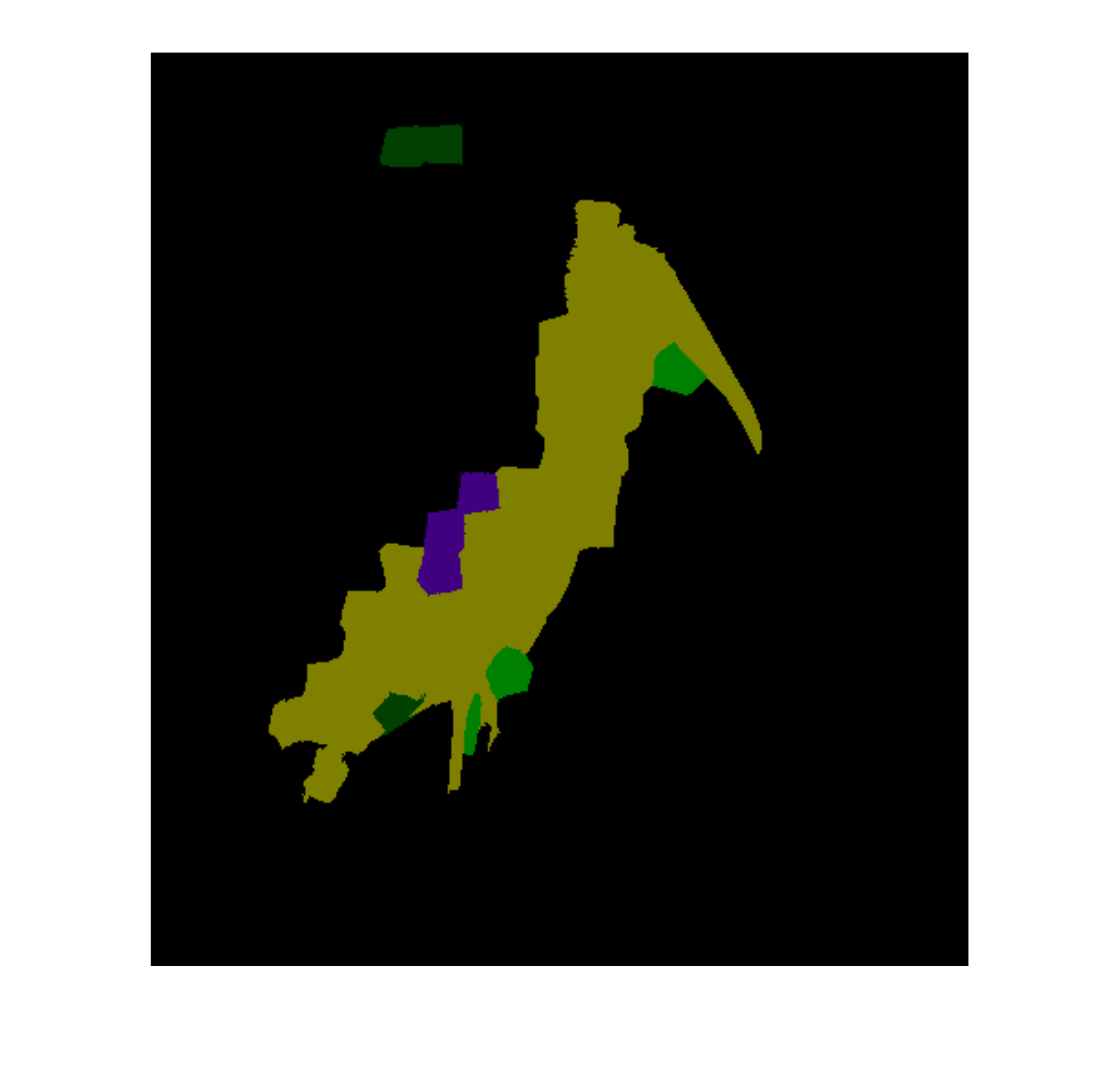} &
    \includegraphics[width=.2\textwidth]{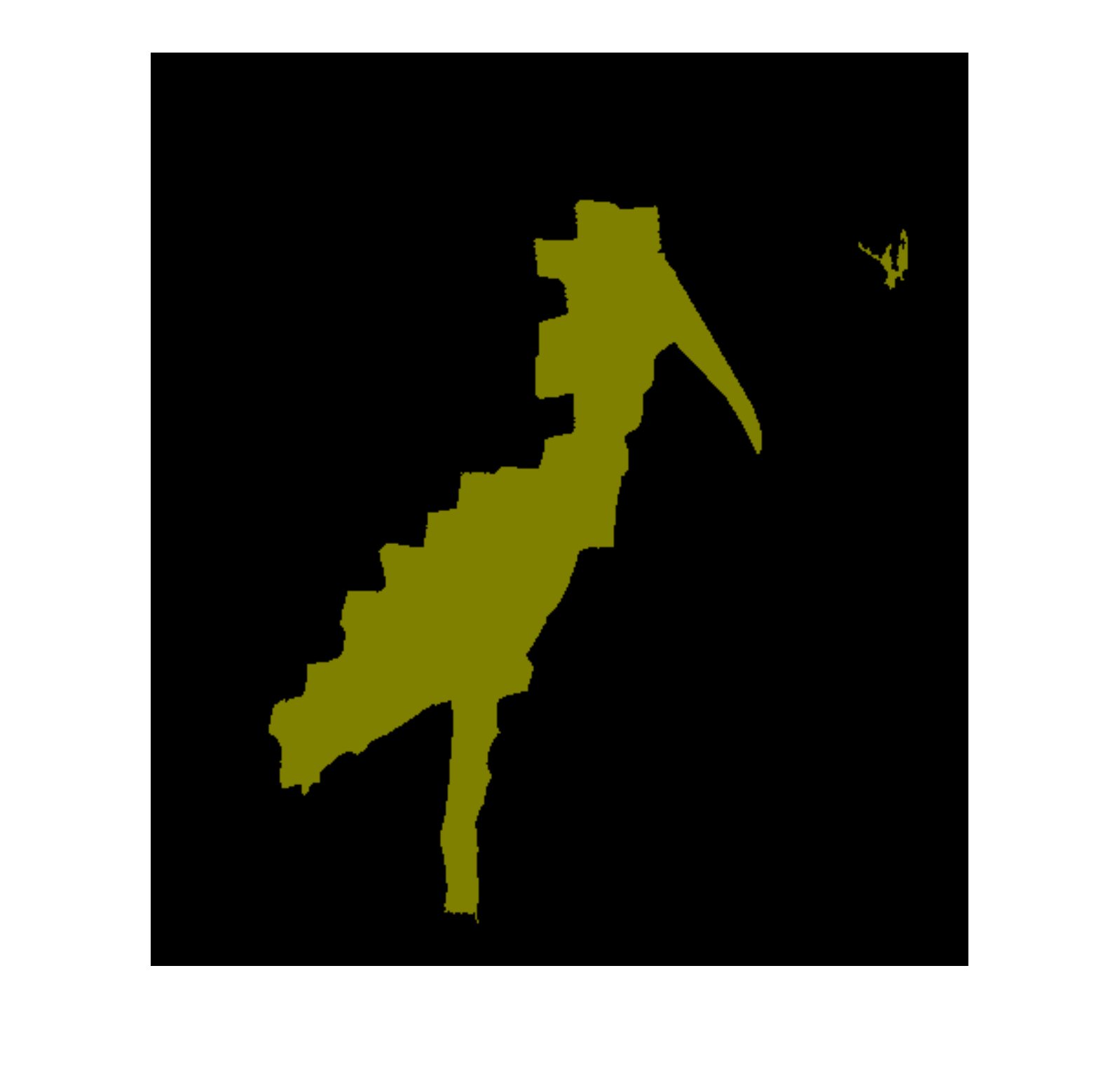} &
    \includegraphics[width=.2\textwidth]{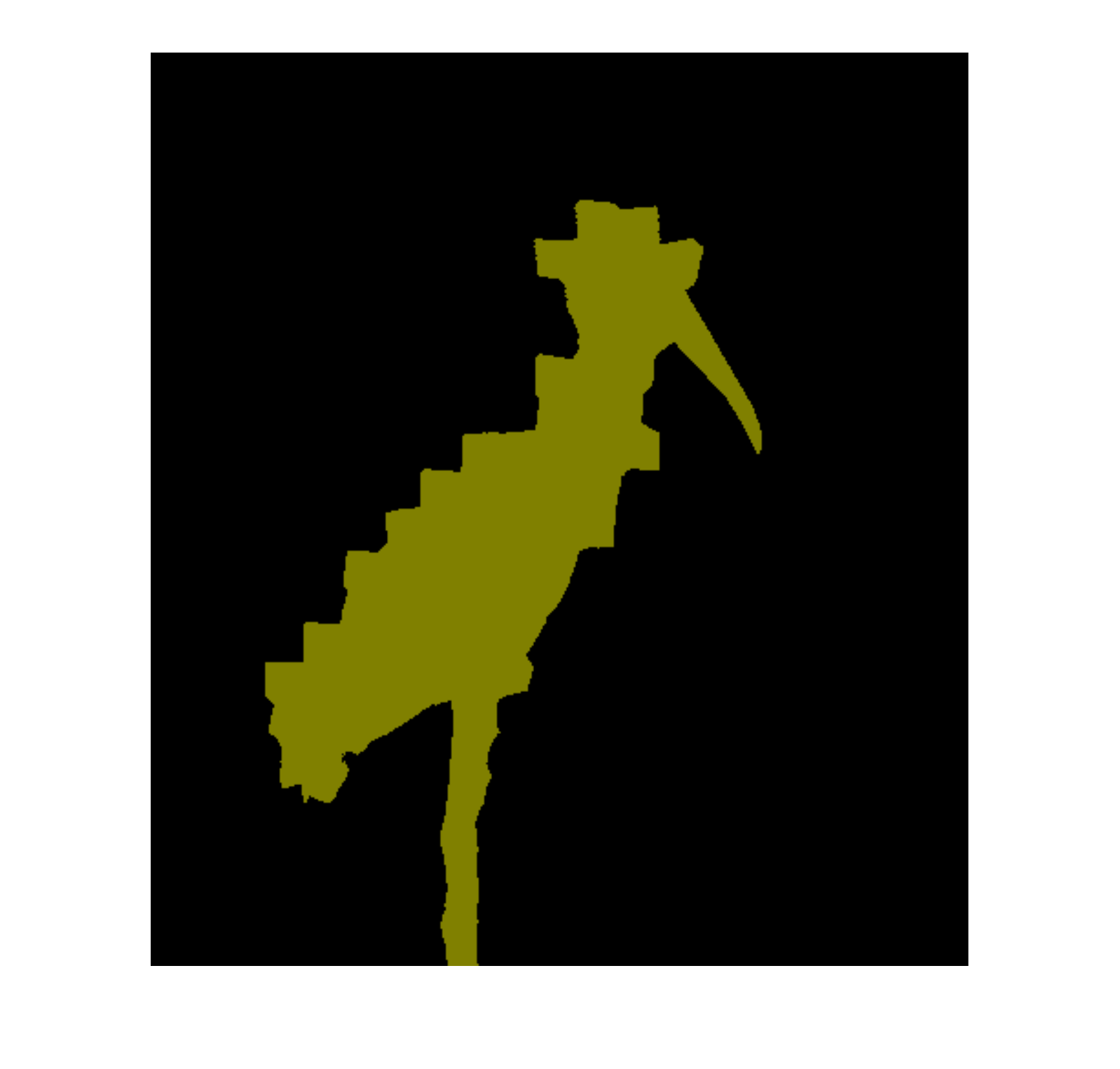}
  \end{tabular}
  \caption{Examples illustrating the effect of zoom-out levels. From
    left: original image, ground truth, local only, local and
    proximal, local, proximal and global, and the full (four levels)
    set of zoom-out features. In all cases a linear model is used to
    label superpixels. See Figure~\ref{fig:colorcode} for category
    color code.}
  \label{fig:levels}
\end{figure*}

It is evident that each of the zoom-out
levels contributes to the eventual accuracy. The most striking is the
effect of adding the distant and global features computed by convnets;
but local and proximal features are very important as well, and
without those we observe poor segment localization.
We also confirmed empirically that learning with asymmetric loss leads to
dramatically better performance, as shown in
Table~\ref{tab:ablation}, with a few examples in
Figure~\ref{fig:levels}.

Next we explored the impact of switching from linear softmax models to
multilayer neural networks, as well as the effect of switching from
the 7-layer CNN-S to VGG-16.
Results of this set of experiments (mean IU when testing on {\tt val})
are
shown in Table~\ref{tab:models}. The best performer in the set of
experiments using CNN-S was
a two-layer network with 512 hidden units with rectified linear unit
activations. Deeper networks tended to
generalize less well, even when using
dropout~\cite{krizhevsky2012imagenet} in training. Then, switching to
VGG-16, we explored classifiers with this architecture, varying the
number of hidden units. The model with 1024 units in the hidden layer
led to the best accuracy on {\tt val}.

\begin{table}[!th]
  \centering
{\small
  \begin{tabular}{|l|l||l|l|}
    \hline
net & \#layers & \#units & mean IoU\\
\hline
CNN-S &1 (linear) & 0 & 52.4\\
\hline
CNN-S &2 & 256  & 57.9\\
CNN-S &2 & 512 & 59.1\\
CNN-S &3 & 256+256 & 56.4\\
CNN-S &3 & 256+256(dropout) &57.0\\
\hline
VGG-16 & 2 & 256 & 62.3 \\
VGG-16 & 2 & 512 & 63.0\\
VGG-16 & 2 & 1024 & {\bf 63.5}\\
\hline
  \end{tabular}}
  \caption{Results on VOC 2012 {\tt val} with different classification
    models. Net: convnet used to extract proximal and global features:
    CNN-S is the 7-layer network from~\cite{simonyan2014return},
    VGG-16 is the 16-layer network from~\cite{simonyan2014very}.
    Both convnets were pre-trained on ImageNet and used with no fine-tuning for
    our task. \#layers: number of layers in the classifier
    network. \#units: number of units in the hidden layer of the
    classifier network. }
  \label{tab:models}
\end{table}

Based on the results in Table~\ref{tab:models} we took the 2-layer
network with 1024 hidden units, with distant and global features
extracted using VGG-16, as the model of choice, and evaluated it
on the VOC 2012 test data\footnote{The best test result with a
  two-layer classifier using CNN-S
  features was mean IU of 58.4}. We report the results of this evaluation in
Table~\ref{tab:voc-old}; to allow for comparison to previously
published results, we also include numbers on VOC 2010 and 2011 test
sets, which are subsets of 2012. The table makes it clear that our
zoom-out architecture achieves accuracies well above those of any of
the previously published methods. Detailed per-class accuracies in
Table~\ref{tab:details-2012} reveal that this superiority is obtained
on 15 out of 20 object categories, some of them (like dogs, cats, or
trains) by a large margin.

\begin{table}[!th]
  \centering
  \begin{tabular}{|>{\small}l||l|l||l|}
    \hline
Method & {\small VOC2010} & {\small VOC2011} & {\small VOC2012}\\
\hline
zoom-out (ours) & {\bf 64.4} & {\bf 64.1} & {\bf 64.4}\\
\hline
FCN-8s~\cite{long2014fully} & -- & 62.7 & 62.2\\ 
Hypercolumns~\cite{hariharan2014hypercolumns} & -- & -- & 59.2\\
DivMbest+convnet~\cite{dhruv-new} & -- & -- & 52.2\\
SDS~\cite{hariharan2014simultaneous} & -- & 52.6 & 51.6\\
DivMbest+rerank~\cite{yadollahpour2013discriminative} & --&--& 48.1\\
Codemaps~\cite{Li2013codemaps} & --&-- &48.3\\
$O2P$~\cite{carreira2012semantic} & -- & 47.6 & 47.8\\
Regions \& parts\cite{arbelaez_cvpr12} & -- & 40.8 & --\\
D-sampling~\cite{lucchi2011spatial} & 33.5 & -- & --\\
Harmony potentials~\cite{boix_ijcv12} & 40.1 & -- & --\\
\hline
  \end{tabular}
  \caption{Results on VOC 2010, 2011 and 2012 test. Mean IoU is shown, see Table~\ref{tab:details-2012} for per-class
    accuracies of the zoom-out method.}
  \label{tab:voc-old}
\end{table}


\renewcommand{\arraystretch}{1.5}
\setlength{\tabcolsep}{3pt}
\begin{table*}[!th]
  \centering
{\small
  \begin{tabu}{|l|c||c|c|c|c|c|c|c|c|c|c|c|c|c|c|c|c|c|c|c|c|c|}
    \hline
    class &
    \rot{90}{\bf mean} &
    \rot{90}{background} &
    \rot{90}{aeroplane} &
    \rot{90}{bycycle}&
    \rot{90}{bird}&
    \rot{90}{boat}&
    \rot{90}{bottle}&
    \rot{90}{bus}&
    \rot{90}{car}&
    \rot{90}{cat}&
    \rot{90}{chair}&
    \rot{90}{cow}&
    \rot{90}{diningtable}&
    \rot{90}{dog}&
    \rot{90}{horse}&
    \rot{90}{motorbike}&
    \rot{90}{person}&
    \rot{90}{potted plant}&
    \rot{90}{sheep}&
    \rot{90}{sofa}&
    \rot{90}{train}&
    \rot{90}{TV/monitor}
    \\
    \hline
    \rowfont{\footnotesize}
    {\normalsize acc} &
    64.4 &
    89.8 &
    {\bf 81.9}& {\bf 35.1} & {\bf 78.2} & {\bf 57.4} & 56.5 &
    {\bf 80.5} & 74.0 & {\bf 79.8} & 22.4 & {\bf 69.6} &
    {\bf 53.7} & {\bf 74.0} & {\bf 76.0} & {\bf 76.6} & 68.8 &
    44.3 & 70.2 & {\bf 40.2} & 68.9 & {\bf 55.3}\\
    \hline
  \end{tabu}
}
\vspace{1em}\caption{Detailed results of our method on VOC 2012 test. Bold indicates best results on a given class of \emph{any} method we are aware of.}
  \label{tab:details-2012}
\end{table*}

\begin{figure*}[!th]
  \centering
  \begin{tabular}{cccccc}
\includegraphics[width=.16\textwidth]{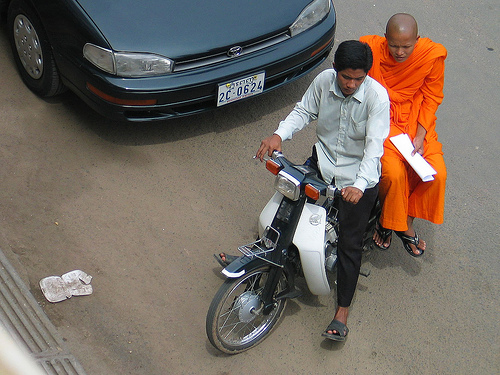} &
 \includegraphics[width=.16\textwidth]{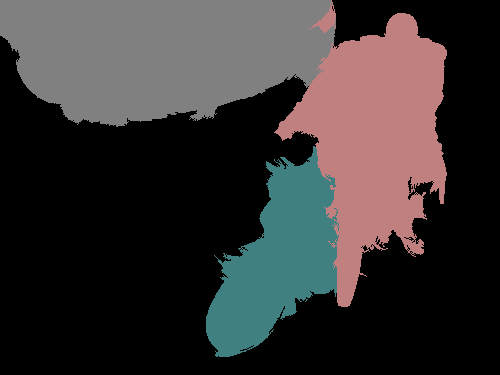}&
\includegraphics[width=.14\textwidth]{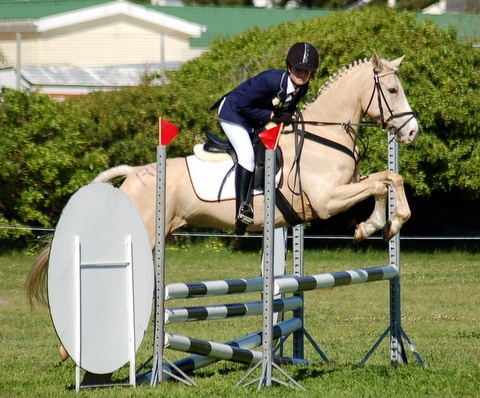} &
 \includegraphics[width=.14\textwidth]{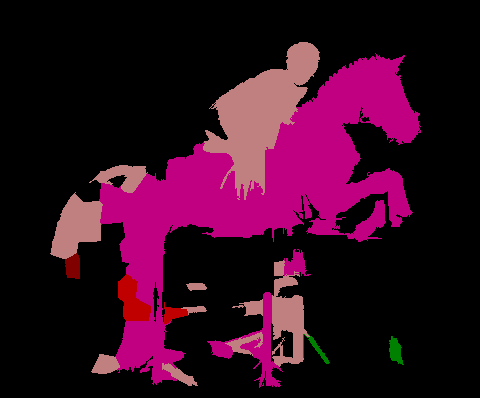}&
\includegraphics[width=.16\textwidth]{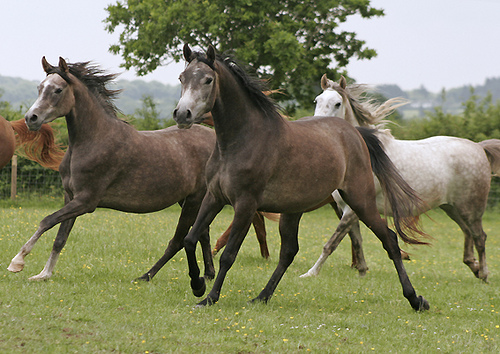} &
 \includegraphics[width=.16\textwidth]{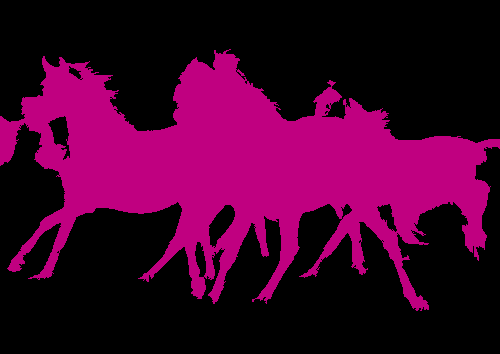}\\
\includegraphics[width=.16\textwidth]{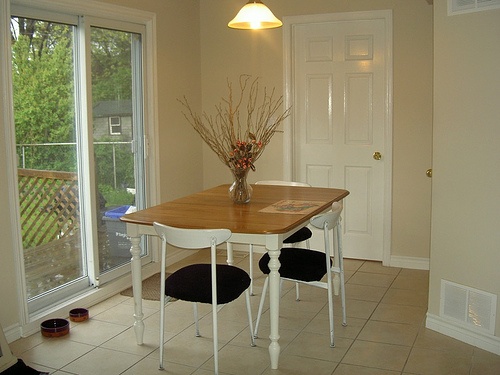} &
 \includegraphics[width=.16\textwidth]{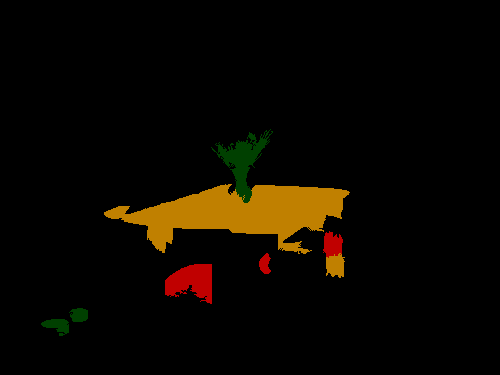}&
\includegraphics[width=.15\textwidth]{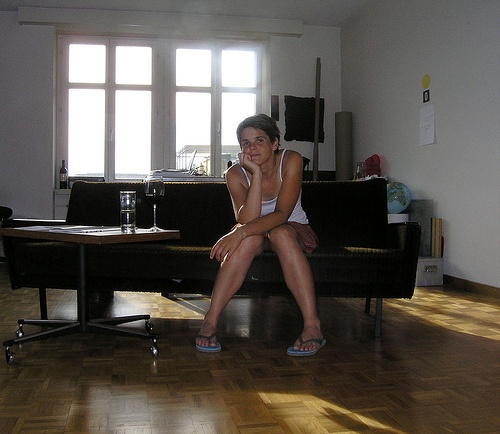} &
 \includegraphics[width=.15\textwidth]{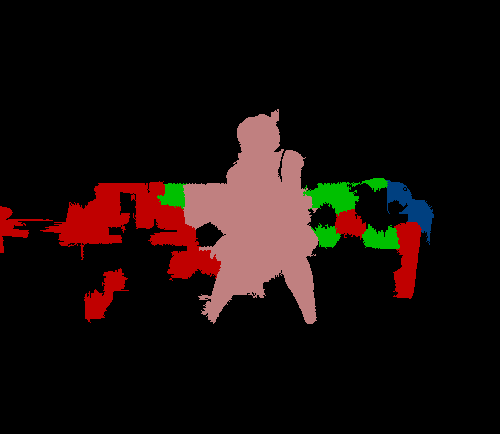}&
\includegraphics[width=.16\textwidth]{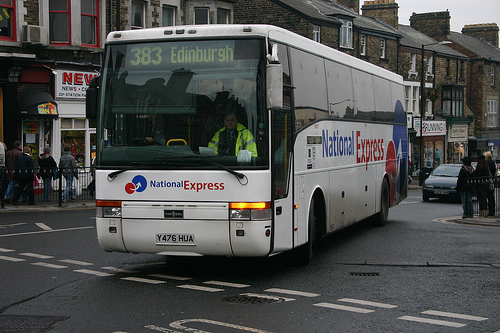} &
 \includegraphics[width=.16\textwidth]{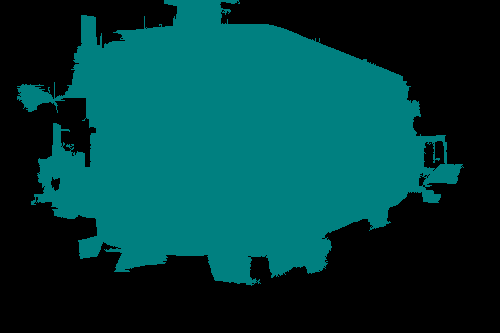}\\
\includegraphics[width=.09\textwidth]{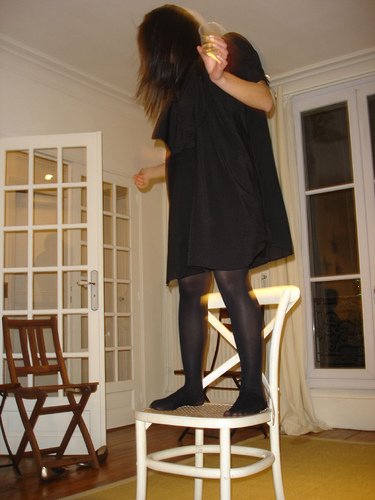} &
 \includegraphics[width=.09\textwidth]{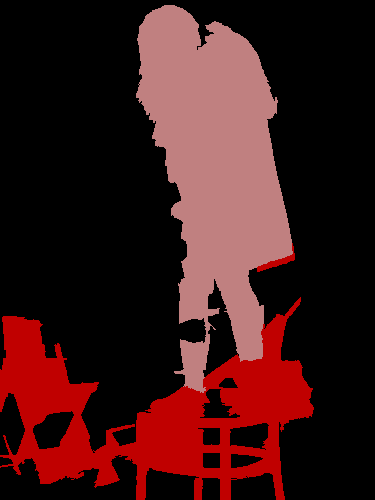} &
\includegraphics[width=.16\textwidth]{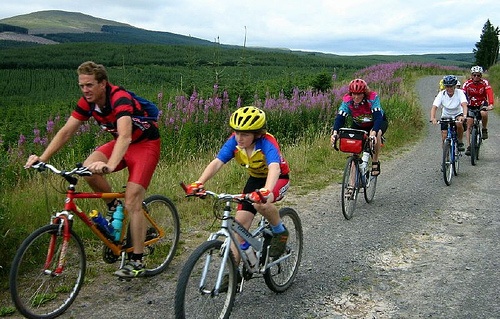} &
 \includegraphics[width=.16\textwidth]{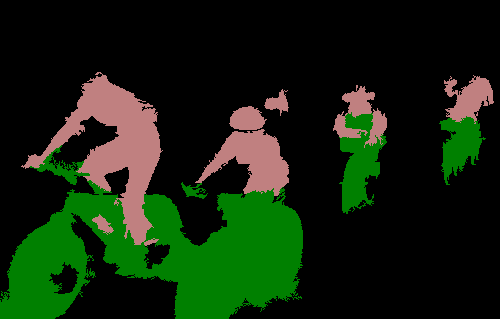}&
\includegraphics[width=.16\textwidth]{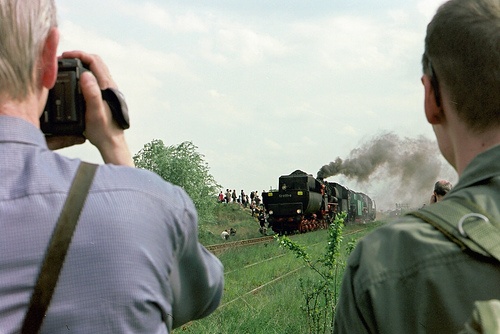} &
 \includegraphics[width=.16\textwidth]{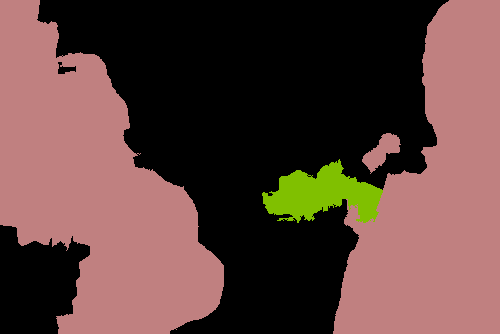}\\
\includegraphics[width=.16\textwidth]{figs/2007_003143.jpg} &
 \includegraphics[width=.16\textwidth]{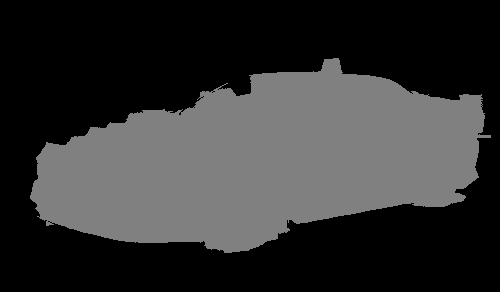}&
\includegraphics[width=.16\textwidth]{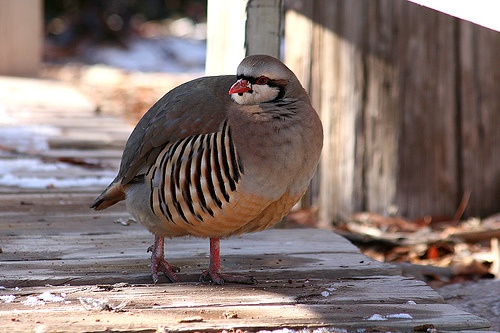} &
 \includegraphics[width=.16\textwidth]{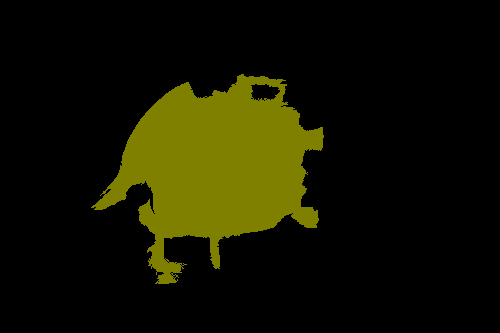}&
\includegraphics[width=.16\textwidth]{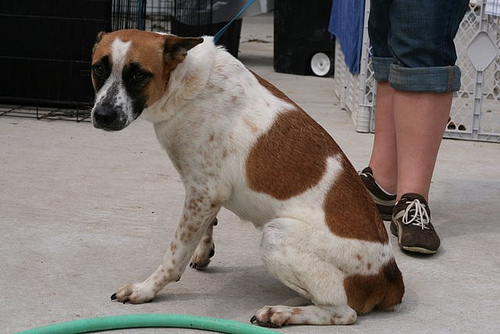} &
 \includegraphics[width=.16\textwidth]{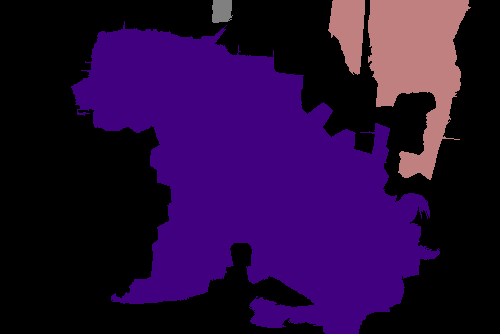}\\
\includegraphics[width=.16\textwidth]{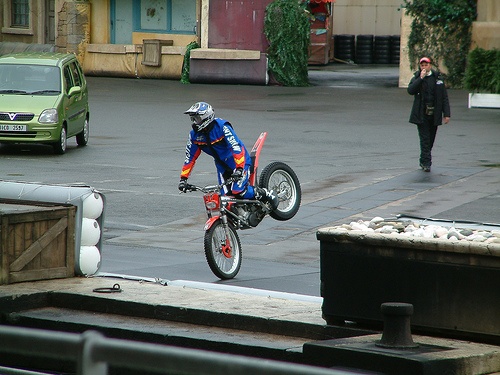} &
 \includegraphics[width=.16\textwidth]{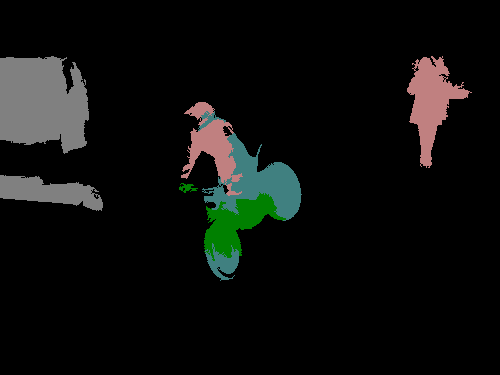}&
\includegraphics[width=.16\textwidth]{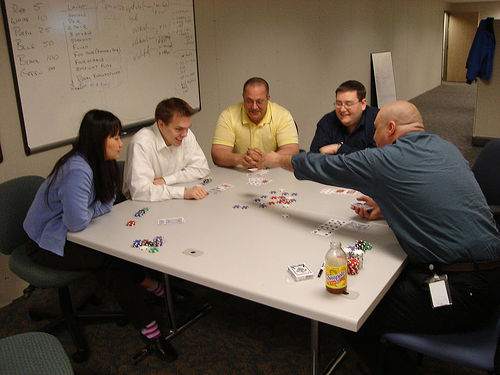} &
 \includegraphics[width=.16\textwidth]{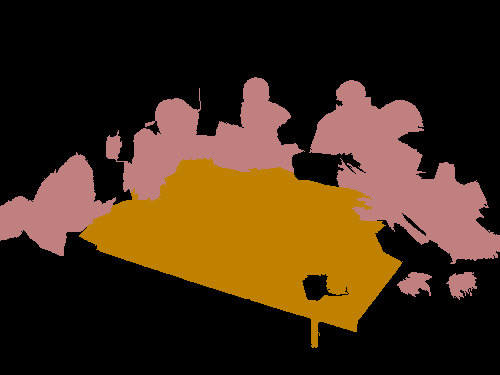}&
\includegraphics[width=.16\textwidth]{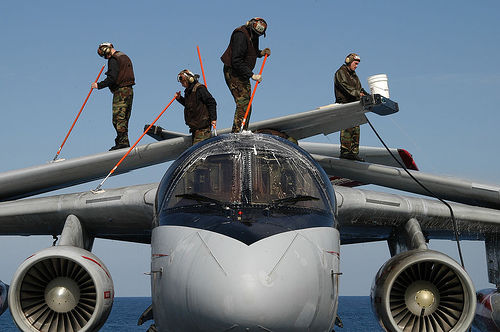} &
 \includegraphics[width=.16\textwidth]{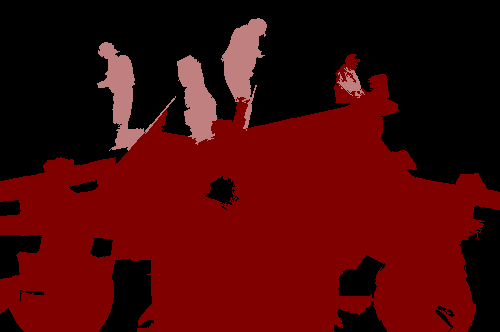}\\
\includegraphics[width=.16\textwidth]{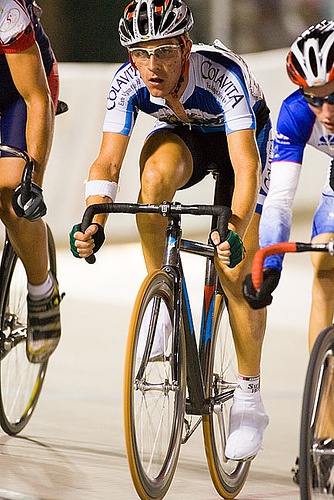} &
 \includegraphics[width=.16\textwidth]{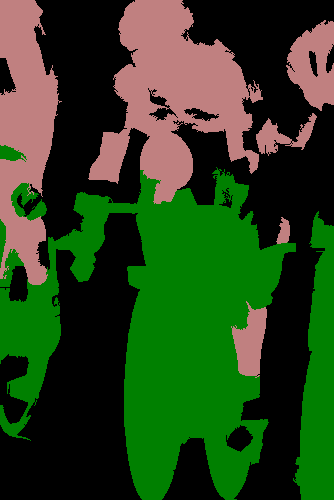}&
\includegraphics[width=.16\textwidth]{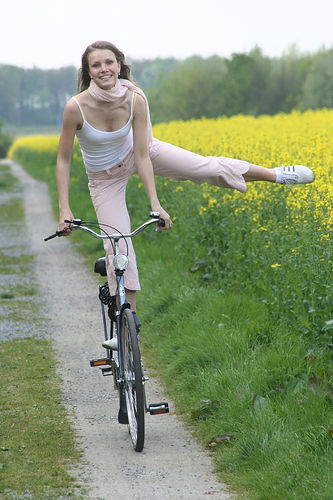} &
 \includegraphics[width=.16\textwidth]{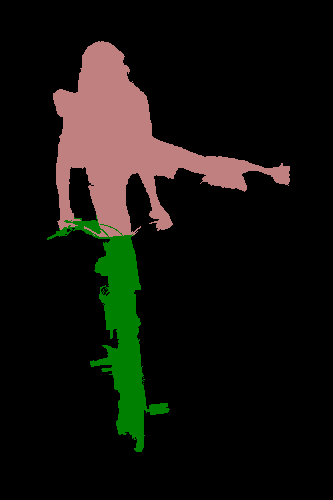}&
\includegraphics[width=.16\textwidth]{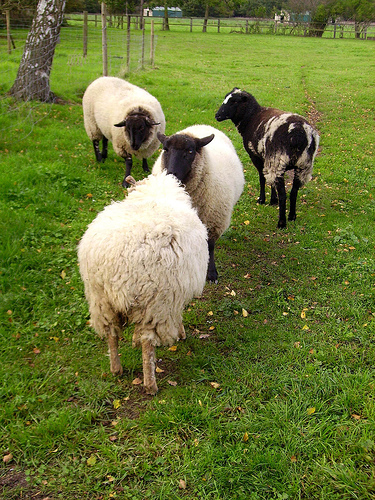} &
 \includegraphics[width=.16\textwidth]{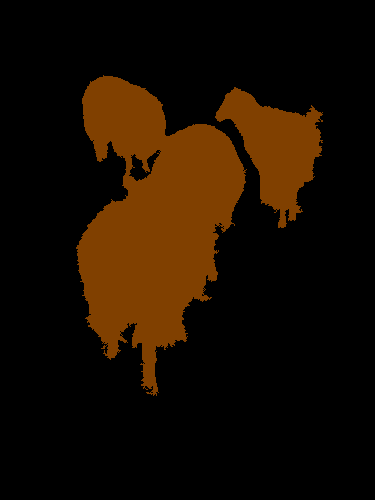}

  \end{tabular}
  \caption{Example segmentations on VOC 2012 val with our best
    classifier (2-layer neural network with 1024 hidden
    units, using four zoom-out levels). See Figure~\ref{fig:colorcode} for category
    color code.}
  \label{fig:examples}
\end{figure*}

\begin{figure*}
  \centering
\begin{tabular}{cccccc}
\includegraphics[width=.16\textwidth]{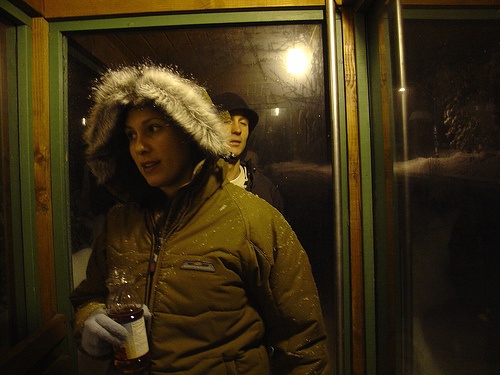} &
 \includegraphics[width=.16\textwidth]{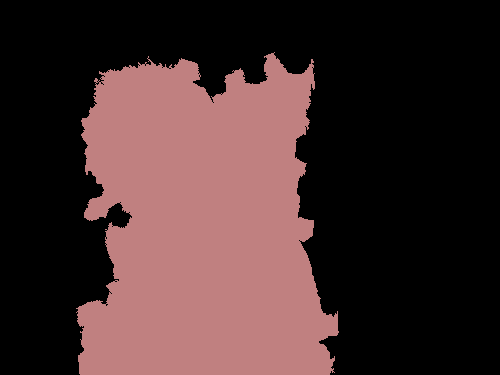}&
 \includegraphics[width=.18\textwidth]{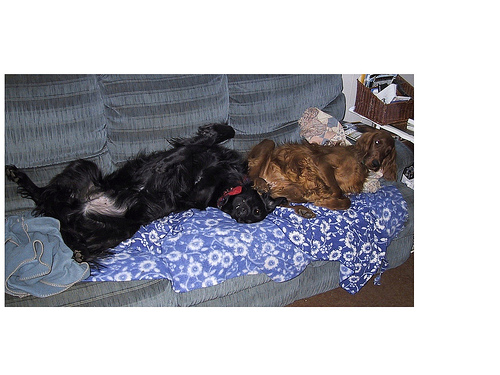} &
 \includegraphics[width=.16\textwidth]{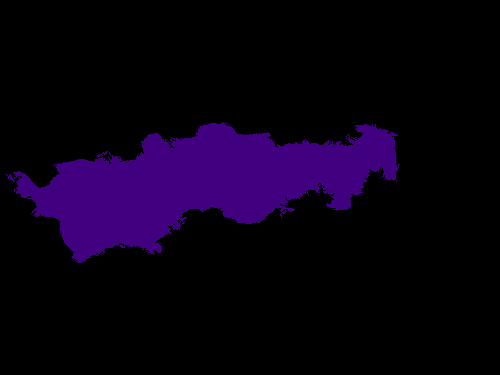}&
\includegraphics[width=.16\textwidth]{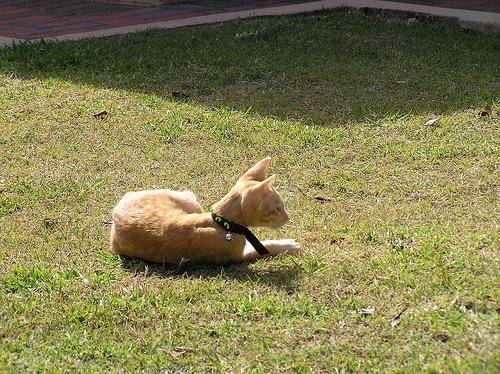} &
 \includegraphics[width=.16\textwidth]{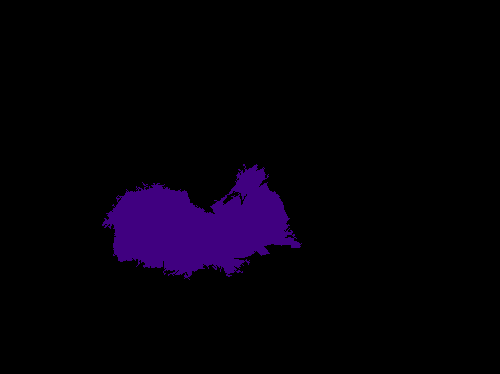}\\
\includegraphics[width=.16\textwidth]{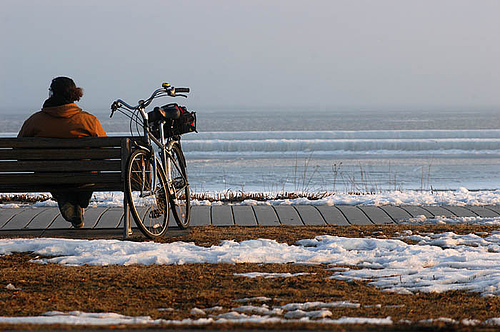} &
 \includegraphics[width=.16\textwidth]{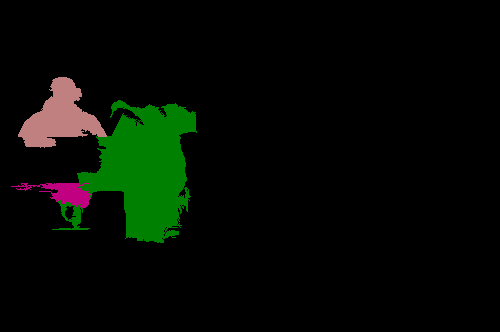}&
\includegraphics[width=.16\textwidth]{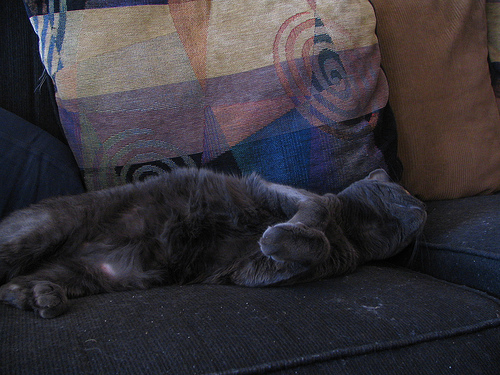} &
 \includegraphics[width=.16\textwidth]{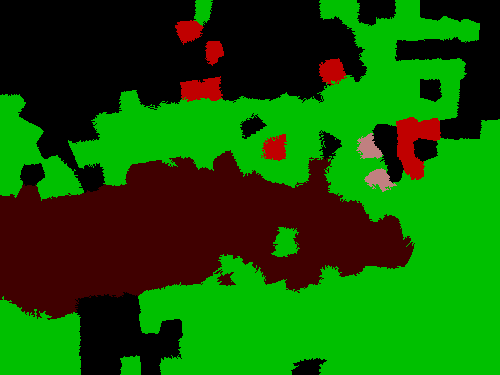}&
\includegraphics[width=.16\textwidth]{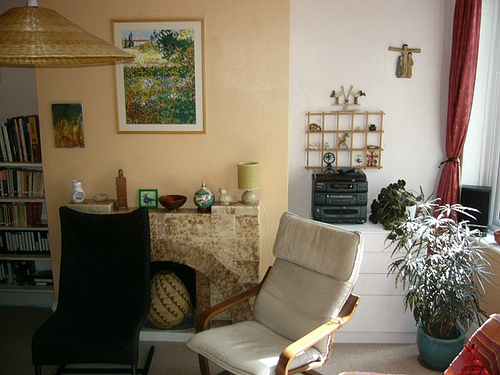} &
 \includegraphics[width=.16\textwidth]{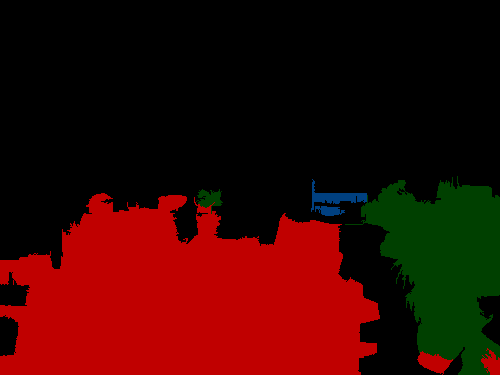}\\
\includegraphics[width=.16\textwidth]{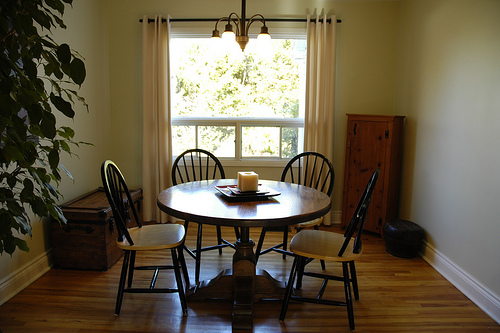} &
 \includegraphics[width=.16\textwidth]{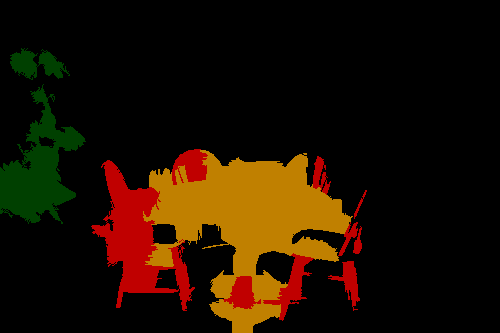}&
\includegraphics[width=.16\textwidth]{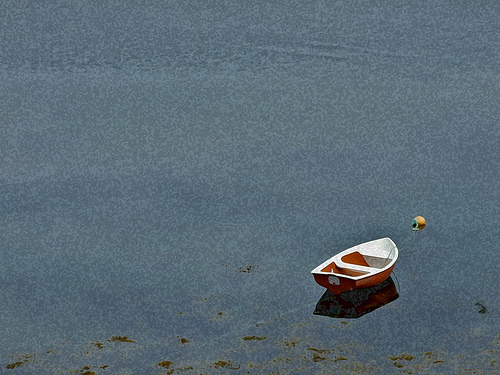} &
 \includegraphics[width=.16\textwidth]{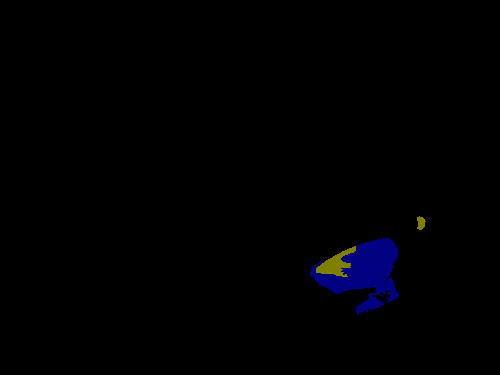}&
\includegraphics[width=.16\textwidth]{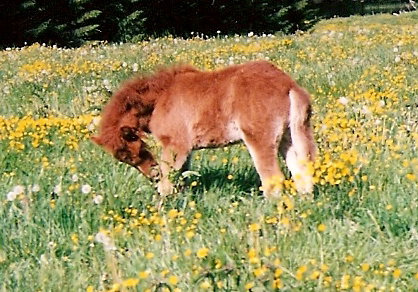} &
 \includegraphics[width=.16\textwidth]{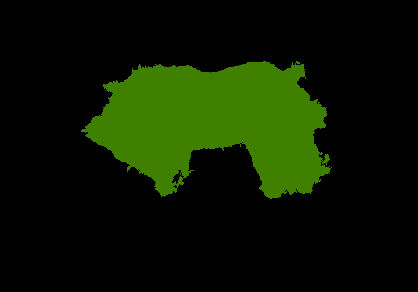}\\
\includegraphics[width=.16\textwidth]{figs/2009_000964.jpg} &
 \includegraphics[width=.16\textwidth]{figs/2009_000964.png}&
\includegraphics[width=.16\textwidth]{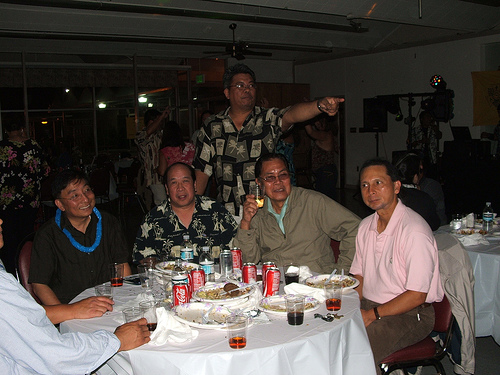} &
 \includegraphics[width=.16\textwidth]{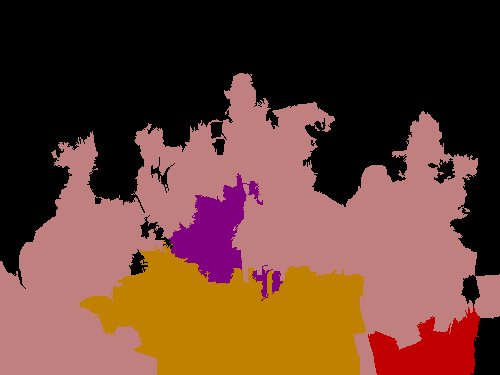}&
\includegraphics[width=.16\textwidth]{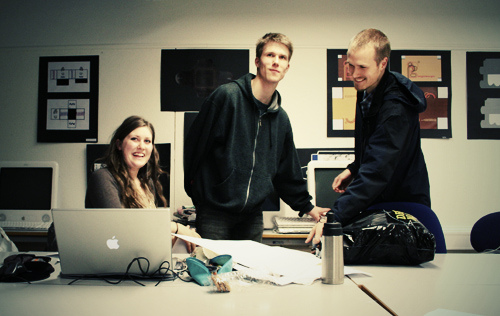} &
 \includegraphics[width=.16\textwidth]{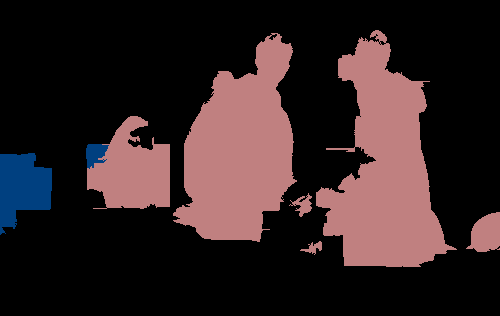}\\
\includegraphics[width=.16\textwidth]{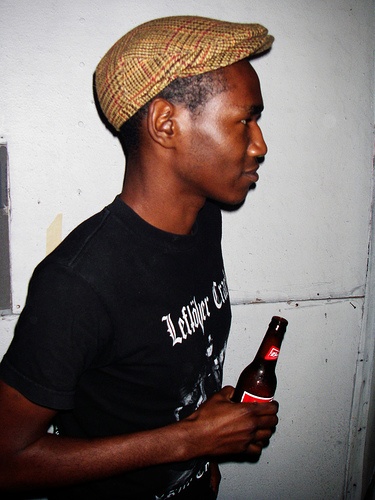} &
 \includegraphics[width=.16\textwidth]{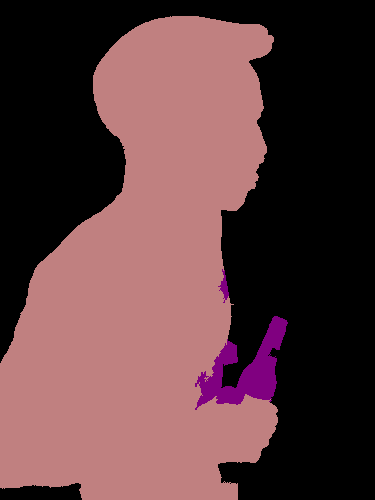}&
\includegraphics[width=.14\textwidth]{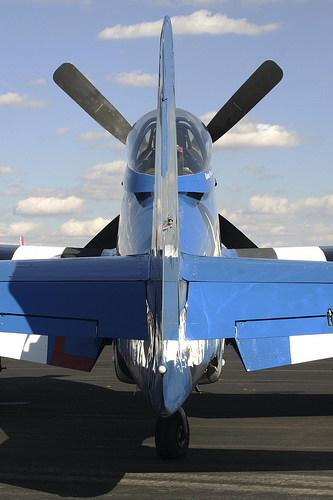} &
 \includegraphics[width=.14\textwidth]{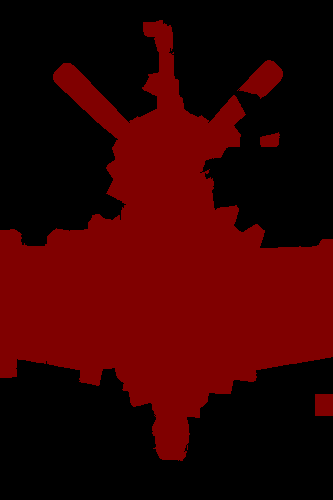}&
\includegraphics[width=.16\textwidth]{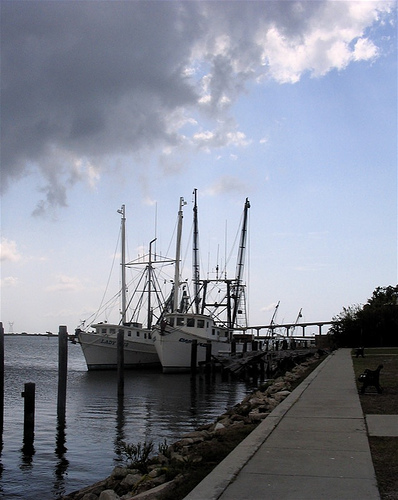}&
 \includegraphics[width=.16\textwidth]{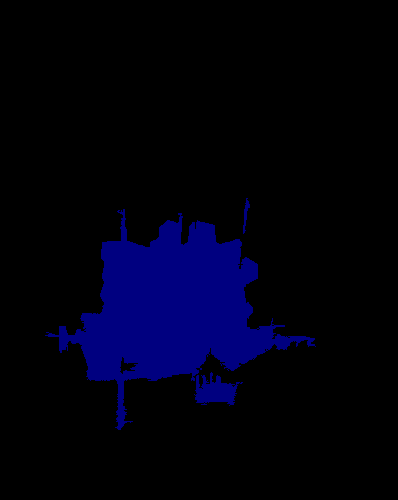}
\end{tabular}
  \caption{Additional examples of segmentation on VOC 2012 val with
    our model. See Figure~\ref{fig:colorcode} for category
    color code.}\label{fig:examples2}  
\end{figure*}

Figure~\ref{fig:examples} displays example
segmentations.
Many of the segmentations have moderate to high accuracy, capturing
correct classes, in correct layout, and sometimes including level of detail
that is usually missing from over-smoothed segmentations obtained by
CRFs or generated by region proposals. But there is tradeoff: despite
the smoothness imposed by higher zoom-out levels, the segmentations we
get do tend to be \emph{under}-smoothed, and in particular include
little ``islands'' of often irrelevant categories. To some extent this
might be alleviated by post-processing; we found that we could learn a
classifier for isolated regions that with reasonable accuracy decides
when these must be ``flipped'' to the surrounding label, and this
improves results on {\tt val} by about 0.5\%, while making the segmentations
more visually pleasing. We do not pursue this ad-hoc approach, and
instead discuss in Section~\ref{sec:conclusions} more principled
remedies that should be investigated in the future.

\subsection{Results on Stanford Background Dataset}
For some of the closely related recent work results on VOC are not
available, so to allow for empirical comparison, we also ran an
experiment on Stanford Background Dataset (SBD). It has 715 images of
outdoor scenes, with
dense labels for eight categories. We applied the same zoom-out
architecture to this dataset as to VOC, with two exceptions: (i) the
local convnet produced 8 features instead of 21+2, and (ii) the
classifier was smaller, with only 128 hidden units, since SBD has about 20 times fewer examples
than VOC and thus could not support training larger models.

There is no standard train/test partition of SBD; the established
protocol calls for reporting 5-fold cross validation results. There is
also no single performance measure; two commonly reported measures are
per-pixel accuracy and average class accuracy (the latter is
different from the VOC measure in that it does not directly penalize false positives).

\begin{table}[!th]
  \centering
{\small
  \begin{tabular}{|l||l|l|}
    \hline
Method & pixel accuracy & class accuracy\\
\hline
zoom-out (ours) & {\bf 82.1} & {\bf 77.3}\\
\hline
Multiscale convnet~\cite{farabet2013learning} & 81.4 & 76.0\\
Recurrent CNN~\cite{pinheiro2014recurrent} & 80.2 & 69.9\\
Pylon~\cite{lempitsky2011pylon} & 81.9 & 72.4 \\
Recursive NN~\cite{socher2011parsing}& 78.1 & -- \\
Multilevel~\cite{mostajabi2014robust} & 78.4 & --\\
\hline
  \end{tabular}
  \caption{Results on Stanford Background Dataset}
  \label{tab:sbd}}
\end{table}

The resuls in Table~\ref{tab:sbd} show that the zoom-out architecture
obtains results better than those in~\cite{pinheiro2014recurrent}
and~\cite{farabet2013learning}, both in class accuracy and in pixel
accuracy.

%% file: colorcode.tex
\begin{figure*}[!th]
  \centering
\begin{tikzpicture}[every node/.style={font=\footnotesize}]
  \definecolor{aeroplane-co}{rgb}{0.502,0,0}
    \node[text=white,fill=aeroplane-co,draw=aeroplane-co,minimum
    height=.5cm,minimum width=1.65cm] (aeroplane) {aeroplane};

    \definecolor{bicycle-co}{rgb}{0,0.502,0}    
    \node[text=white,fill=bicycle-co,draw=bicycle-co,minimum
    height=.5cm,minimum width=1.65cm,right=of aeroplane,right=.1cm] (bicycle) {bicycle};

    \definecolor{bird-co}{rgb}{0.502,0.502,0}    
    \node[text=white,fill=bird-co,draw=bird-co,minimum
    height=.5cm,minimum width=1.65cm,right=of bicycle,right=.1cm] (bird) {bird};

    \definecolor{boat-co}{rgb}{0,0,0.502}    
        \node[text=white,fill=boat-co,draw=boat-co,minimum
    height=.5cm,minimum width=1.65cm,right=of bird,right=.1cm] (boat) {boat};

    \definecolor{bottle-co}{rgb}{0.502,0,0.502}    
    \node[text=white,fill=bottle-co,draw=bottle-co,minimum
    height=.5cm,minimum width=1.65cm,right=of boat,right=.1cm] (bottle) {bottle};
    
    \definecolor{bus-co}{rgb}{0,0.502,0.502}    
    \node[text=white,fill=bus-co,draw=bus-co,minimum
    height=.5cm,minimum width=1.65cm,right=of bottle,right=.1cm] (bus) {bus};

    \definecolor{car-co}{rgb}{0.502,0.502,0.502}    
    \node[text=white,fill=car-co,draw=car-co,minimum
    height=.5cm,minimum width=1.65cm,right=of bus,right=.1cm] (car)
    {car};

    \definecolor{cat-co}{rgb}{0.251,0,0}    
    \node[text=white,fill=cat-co,draw=cat-co,minimum
    height=.5cm,minimum width=1.65cm,right=of car,right=.1cm] (cat)
    {cat};

    \definecolor{chair-co}{rgb}{0.753,0,0}    
    \node[text=white,fill=chair-co,draw=chair-co,minimum
    height=.5cm,minimum width=1.65cm,right=of cat,right=.1cm] (chair)
    {chair};

    \definecolor{cow-co}{rgb}{0.251,0.502,0}    
    \node[text=white,fill=cow-co,draw=cow-co,minimum
    height=.5cm,minimum width=1.65cm,right=of chair,right=.1cm] (cow)
    {cow};

    \definecolor{diningtable-co}{rgb}{0.753,0.502,0}    
    \node[text=white,fill=diningtable-co,draw=diningtable-co,minimum
    height=.5cm,minimum width=1.65cm,below=of aeroplane,below=.1cm] (diningtable)
    {diningtable};

    \definecolor{dog-co}{rgb}{0.251,0,0.502}    
    \node[text=white,fill=dog-co,draw=dog-co,minimum
    height=.5cm,minimum width=1.65cm,right=of diningtable,right=.1cm] (dog)
    {dog};

    \definecolor{horse-co}{rgb}{0.753,0,0.502}    
    \node[text=white,fill=horse-co,draw=horse-co,minimum
    height=.5cm,minimum width=1.65cm,right=of dog,right=.1cm] (horse)
    {horse};

    \definecolor{motorbike-co}{rgb}{0.251,0.502,0.502}    
    \node[text=white,fill=motorbike-co,draw=motorbike-co,minimum
    height=.5cm,minimum width=1.65cm,right=of horse,right=.1cm] (motorbike)
    {motorbike};

    \definecolor{person-co}{rgb}{0.753,0.502,0.502}    
    \node[text=white,fill=person-co,draw=person-co,minimum
    height=.5cm,minimum width=1.65cm,right=of motorbike,right=.1cm] (person)
    {person};

    \definecolor{pottedplant-co}{rgb}{0,0.251,0}    
    \node[text=white,fill=pottedplant-co,draw=pottedplant-co,minimum
    height=.5cm,minimum width=1.65cm,right=of person,right=.1cm] (pottedplant)
    {pottedplant};

    \definecolor{sheep-co}{rgb}{0.502,0.251,0}    
    \node[text=white,fill=sheep-co,draw=sheep-co,minimum
    height=.5cm,minimum width=1.65cm,right=of pottedplant,right=.1cm] (sheep)
    {sheep};

    \definecolor{sofa-co}{rgb}{0,0.753,0}    
    \node[text=white,fill=sofa-co,draw=sofa-co,minimum
    height=.5cm,minimum width=1.65cm,right=of sheep,right=.1cm] (sofa)
    {sofa};

    \definecolor{train-co}{rgb}{0.502,0.753,0}    
    \node[text=white,fill=train-co,draw=train-co,minimum
    height=.5cm,minimum width=1.65cm,right=of sofa,right=.1cm] (train)
    {train};

   \definecolor{tvmonitor-co}{rgb}{0,0.251,0.502}    
    \node[text=white,fill=tvmonitor-co,draw=tvmonitor-co,minimum
    height=.5cm,minimum width=1.65cm,right=of train,right=.1cm] (tvmonitor)
    {tvmonitor};

\end{tikzpicture}
  \caption{Color code for VOC categories. Background is black.}
  \label{fig:colorcode}
\end{figure*}

%% file: conclusions.tex
\section{Conclusions}\label{sec:conclusions}

The main point of this paper is to explore how far we can push
feedforward semantic labeling of superpixels when we use
multilevel, zoom-out feature construction and train non-linear
classifiers (multi-layer neural networks) with asymmetric loss. The results are perhaps surprising:
we can far surpass existing state of the art, despite apparent
simplicity of our method and lack of explicit respresentation of the
structured nature of the segmentation task. Another important
conclusion that emerges from this is that we finally have shown that
segmentation, just like image classification, detection and other
recognition tasks, can benefit from the advent of deep convolutional
networks. 

Despite this progress, much remains to be
done as we continue this work. Immediately, we plan to
explore obvious improvements. One of these is replacement of
all the handcrafted local and proximal features with features learned
from the data with convnets. Another such improvement is to fine tune
the ``off the shelf'' networks currently used for distant and global
feature extraction. We also plan to look into including additional
zoom-out levels.
Our longer term plan is to switch from a collection of feature
extractors deployed at different zoom-out levels to a single zoom-out
network, which would be more in line with the philosophy of end-to-end
learning that has drive deep learning research.

Finally, despite the surprising success of the
zoom-out architecture described here, we by no means intend to entirely dismiss CRFs,
or more generally inference in structured models. We believe that the
zoom-out architecture may eventually benefit from
bringing back some form of inference to ``clean up'' the predictions. We hope that this can be done
without giving up the feedforward nature of our approach; one
possibility we are interested in exploring is to ``unroll''
approximate inference
into additional layers in the feedforward network~\cite{li2014mean,stoyanov2011empirical}.

%% file: arxiv.bbl
\begin{thebibliography}{10}\itemsep=-1pt

\bibitem{achanta_pami12}
R.~Achanta, A.~Shaji, K.~Smith, A.~Lucchi, P.~Fua, and S.~SŸsstrunk.
\newblock Slic superpixels compared to state-of-the-art superpixel methods.
\newblock {\em IEEE TPAMI}, 2012.

\bibitem{arbelaez_cvpr12}
P.~Arbelaez, B.~Hariharan, C.~Gu, S.~Gupta, L.~Bourdev, and J.~Malik.
\newblock Semantic segmentation using regions and parts.
\newblock In {\em CVPR}, 2012.

\bibitem{boix_ijcv12}
X.~Boix, J.~M. Gonfaus, J.~van~de Weijer, A.~D. Bagdanov, J.~S. Gual, and
  J.~Gonz{\`a}lez.
\newblock Harmony potentials - fusing global and local scale for semantic image
  segmentation.
\newblock {\em IJCV}, 96(1):83--102, 2012.

\bibitem{carreira2012semantic}
J.~Carreira, R.~Caseiro, J.~Batista, and C.~Sminchisescu.
\newblock Semantic segmentation with second-order pooling.
\newblock In {\em Computer Vision--ECCV 2012}, pages 430--443. Springer, 2012.

\bibitem{carreira_ijcv12}
J.~Carreira, F.~Li, and C.~Sminchisescu.
\newblock Object recognition by sequential figure-ground ranking.
\newblock {\em IJCV}, 98(3):243--262, 2012.

\bibitem{carreira2012cpmc}
J.~Carreira and C.~Sminchisescu.
\newblock Cpmc: Automatic object segmentation using constrained parametric
  min-cuts.
\newblock {\em Pattern Analysis and Machine Intelligence, IEEE Transactions
  on}, 34(7), 2012.

\bibitem{simonyan2014return}
K.~Chatfield, K.~Simonyan, A.~Vedaldi, and A.~Zisserman.
\newblock Return of the devil in the details: Delving deep into convolutional
  nets.
\newblock In {\em BMVC}, 2014.

\bibitem{dhruv-new}
M.~Cogswell, X.~Lin, and D.~Batra.
\newblock Personal communication.
\newblock November 2014.

\bibitem{everingham2014pascal}
M.~Everingham, S.~Eslami, L.~Van~Gool, C.~Williams, J.~Winn, and A.~Zisserman.
\newblock The pascal visual object classes challenge: A retrospective.
\newblock {\em International Journal of Computer Vision}, 2014.

\bibitem{farabet2013learning}
C.~Farabet, C.~Couprie, L.~Najman, and Y.~LeCun.
\newblock Learning hierarchical features for scene labeling.
\newblock {\em IEEE TPAMI}, 35(8), 2013.

\bibitem{fulkerson2009class}
B.~Fulkerson, A.~Vedaldi, and S.~Soatto.
\newblock Class segmentation and object localization with superpixel
  neighborhoods.
\newblock In {\em CVPR}, 2009.

\bibitem{girshick2014rich}
R.~Girshick, J.~Donohue, T.~Darrell, and J.~Malik.
\newblock Rich feature hierarchies for accurate object detection and semantic
  segmentation.
\newblock {\em arXiv preprint http://arxiv.org/abs/1311.2524}, 2014.

\bibitem{hariharan2014hypercolumns}
B.~Hariharan, P.~A. an~R.~Girshick, and J.~Malik.
\newblock Hypercolumns for object segmentation and fine-grained localization.
\newblock {\em arXiv preprint http://arxiv.org/abs/1411.5752}, 2014.

\bibitem{BharathICCV2011}
B.~Hariharan, P.~Arbelaez, L.~Bourdev, S.~Maji, and J.~Malik.
\newblock Semantic contours from inverse detectors.
\newblock In {\em International Conference on Computer Vision (ICCV)}, 2011.

\bibitem{hariharan2014simultaneous}
B.~Hariharan, P.~Arbel{\'a}ez, R.~Girshick, and J.~Malik.
\newblock Simultaneous detection and segmentation.
\newblock In {\em Computer Vision--ECCV 2014}, 2014.

\bibitem{huang2013deep}
G.~B. Huang and V.~Jain.
\newblock Deep and wide multiscale recursive networks for robust image
  labeling.
\newblock {\em arXiv preprint arXiv:1310.0354}, 2013.

\bibitem{ion2011probabilistic}
A.~Ion, J.~Carreira, and C.~Sminchisescu.
\newblock Probabilistic joint image segmentation and labeling.
\newblock In {\em NIPS}, pages 1827--1835, 2011.

\bibitem{jia2014caffe}
Y.~Jia, E.~Shelhamer, J.~Donahue, S.~Karayev, J.~Long, R.~Girshick,
  S.~Guadarrama, and T.~Darrell.
\newblock Caffe: Convolutional architecture for fast feature embedding.
\newblock {\em arXiv preprint arXiv:1408.5093}, 2014.

\bibitem{krizhevsky2012imagenet}
A.~Krizhevsky, I.~Sutskever, and G.~E. Hinton.
\newblock Imagenet classification with deep convolutional neural networks.
\newblock In {\em NIPS}, 2012.

\bibitem{kuettel2012segmentation}
D.~Kuettel, M.~Guillaumin, and V.~Ferrari.
\newblock Segmentation propagation in imagenet.
\newblock In {\em ECCV}, 2012.

\bibitem{ladicky_iccv09}
L.~Ladick{\` y}, C.~Russell, P.~Kohli, and P.~H.~S. Torr.
\newblock Associative hierarchical {CRF}s for object class image segmentation.
\newblock {\em ICCV}, 2009.

\bibitem{lempitsky2011pylon}
V.~Lempitsky, A.~Vedaldi, and A.~Zisserman.
\newblock Pylon model for semantic segmentation.
\newblock In {\em NIPS}, pages 1485--1493, 2011.

\bibitem{li2014mean}
Y.~Li and R.~Zemel.
\newblock Mean field networks.
\newblock In {\em ICML Workshop on Learning Tractable Probabilistic Models},
  2014.

\bibitem{Li2013codemaps}
Z.~Li, E.~Gavves, K.~E.~A. van~de Sande, C.~G.~M. Snoek, and A.~W.~M.
  Smeulders.
\newblock Codemaps segment, classify and search objects locally.
\newblock In {\em ICCV}, 2013.

\bibitem{lim2009context}
J.~J. Lim, P.~Arbel{\'a}ez, C.~Gu, and J.~Malik.
\newblock Context by region ancestry.
\newblock In {\em ICCV}, 2009.

\bibitem{long2014fully}
J.~Long, E.~Shelhamer, and T.~Darrell.
\newblock Fully convolutional networks for semantic segmentation.
\newblock {\em arXiv preprint http://arxiv.org/abs/1411.4038}, 2014.

\bibitem{lucchi2011spatial}
A.~Lucchi, Y.~Li, X.~Boix, K.~Smith, and P.~Fua.
\newblock Are spatial and global constraints really necessary for segmentation?
\newblock In {\em ICCV}, 2011.

\bibitem{mostajabi2014robust}
M.~Mostajabi and I.~Gholampour.
\newblock A robust multilevel segment description for multi-class object
  recognition.
\newblock {\em Machine Vision and Applications}, 2014.

\bibitem{pinheiro2014recurrent}
P.~H.~O. Pinheiro and R.~Collobert.
\newblock Recurrent convolutional neural networks for scene labeling.
\newblock In {\em ICML}, 2014.

\bibitem{russakovsky2014imagenet}
O.~Russakovsky, J.~Deng, H.~Su, J.~Krause, S.~Satheesh, S.~Ma, Z.~Huang,
  A.~Karpathy, A.~Khosla, M.~Bernstein, et~al.
\newblock Imagenet large scale visual recognition challenge.
\newblock {\em arXiv:1409.0575}, 2014.

\bibitem{shotton_ijcv09}
J.~Shotton, J.~Winn, C.~Rother, and A.~Criminisi.
\newblock Textonboost for image understanding: Multi-class object recognition
  and segmentation by jointly modeling texture, layout, and context.
\newblock {\em IJCV}, 81(1), 2009.

\bibitem{simonyan2014very}
K.~Simonyan and A.~Zisserman.
\newblock Very deep convolutional networks for large-scale image recognition.
\newblock {\em arXiv preprint http://arxiv.org/abs/1409.1556}, 2014.

\bibitem{socher2011parsing}
R.~Socher, C.~C. Lin, A.~Y. Ng, and C.~D. Manning.
\newblock {Parsing Natural Scenes and Natural Language with Recursive Neural
  Networks}.
\newblock In {\em ICML}, 2011.

\bibitem{stoyanov2011empirical}
V.~Stoyanov, A.~Ropson, and J.~Eisner.
\newblock Empirical risk minimization of graphical model parameters given
  approximate inference, decoding, and model structure.
\newblock In {\em AISTATS}, 2011.

\bibitem{tarlow2012structured}
D.~Tarlow and R.~S. Zemel.
\newblock Structured output learning with high order loss functions.
\newblock In {\em AISTATS}, 2012.

\bibitem{UijlingsIJCV2013}
J.~R.~R. Uijlings, K.~E.~A. van~de Sande, T.~Gevers, and A.~W.~M. Smeulders.
\newblock Selective search for object recognition.
\newblock {\em International Journal of Computer Vision}, 104(2), 2013.

\bibitem{yadollahpour2013discriminative}
P.~Yadollahpour, D.~Batra, and G.~Shakhnarovich.
\newblock Discriminative re-ranking of diverse segmentations.
\newblock In {\em CVPR}, 2013.

\end{thebibliography}
